\documentclass[journal,usenames,dvipsnames]{IEEEtran}
\IEEEoverridecommandlockouts
\usepackage{cite}
\usepackage{amsmath,amssymb,amsfonts}
\usepackage{graphicx}
\usepackage{textcomp}
\usepackage{xcolor}

\usepackage{bbm}
\usepackage{booktabs}
\usepackage[table]{xcolor}
\usepackage{makecell}

\usepackage{algorithm,algorithmicx,listings}
\usepackage[noend]{algpseudocode}
\algnewcommand{\LineComment}[1]{{\Statex \vspace{1ex} \(\triangleright\) #1}}
\DeclareMathOperator{\distop}{dist}
\newcommand{\dist}[2]{\distop\!\left(#1,#2\right)}

\usepackage[
    breaklinks=true,
    colorlinks=true,
    linkcolor=blue,
    citecolor=blue,
    urlcolor=black
]{hyperref}
\usepackage{optidef}

\def\BibTeX{{\rm B\kern-.05em{\sc i\kern-.025em b}\kern-.08em
    T\kern-.1667em\lower.7ex\hbox{E}\kern-.125emX}}
\begin{document}

\title{Contact-Aware Planning and Control of Continuum Robots in Highly Constrained Environments\\
\author{Aedan Mangan$^{1}$, Kehan Long$^{2}$, Ki Myung Brian Lee$^{3}$, Miheer Potdar$^{1}$, Nikolay Atanasov$^{3}$, Tania K. Morimoto$^{1, 4}$
}  
\thanks{
$^{1}$Department of Mechanical and Aerospace Engineering, University of California San Diego, San Diego, CA 92093, USA (email: tamorimoto@ucsd.edu)}
\thanks{
$^{2}$SceniX Inc, New York, NY 10011, USA}
\thanks{
$^{3}$Department of Electrical and Computer Engineering, University of California San Diego, San Diego, CA 92093, USA}
\thanks{
$^{4}$Department of Surgery, University of California San Diego, San Diego, CA 92093, USA }
\thanks{This article has supplementary downloadable material (Supplementary Video S1) available, provided by the authors.
}
}

\maketitle

\begin{abstract}
Continuum robots are well suited for navigating confined and fragile environments, such as vascular or endoluminal anatomy, where contact with surrounding structures is often unavoidable. 
While controlled contact can assist motion, unfavorable contact can degrade controllability, induce kinematic singularities, or introduce safety risks. 
We present a contact-aware planning approach that evaluates contact quality, penalizing hazardous interactions, while permitting benign contact.
The planner produces kinematically feasible trajectories and contact-aware Jacobians which can be used for closed-loop control in hardware experiments.
We validate the approach by testing the integrated system (planning, control, and mechanical design) on anatomical models from patient scans.
The planner generates effective plans for three common anatomical environments, and, 
in all hardware trials, the continuum robot was able to reach the target while avoiding dangerous tip contact (100\% success). 
Mean tracking errors were $1.9 \pm 0.5$~mm, $1.2 \pm 0.1$~mm, and $1.7 \pm 0.2$~mm across the three different environments.
Ablation studies showed that penalizing end-of-continuum-segment (ECS) contact improved manipulability and prevented hardware failures. 
Overall, this work enables reliable, contact-aware navigation in highly constrained environments.

\end{abstract}


\section{Introduction} \label{sec_intro}
Continuum or soft robots have become increasingly popular in applications, such as vascular intervention, endoscopy, and inspection in confined spaces \cite{burgner2015continuum, dupont2022continuum, russo2023continuum}, which require navigating through tight, deformable, and fragile environments. In these settings, contact between the robot and the environment is not merely a disturbance to avoid but often a necessary part of successful motion. However, while controlled contact can facilitate navigation, unfavorable contact can degrade controllability, induce kinematic singularities, or cause irreversible damage to surrounding structures.
\begin{figure}[t!]
\centerline{\includegraphics[width=1.0\linewidth]{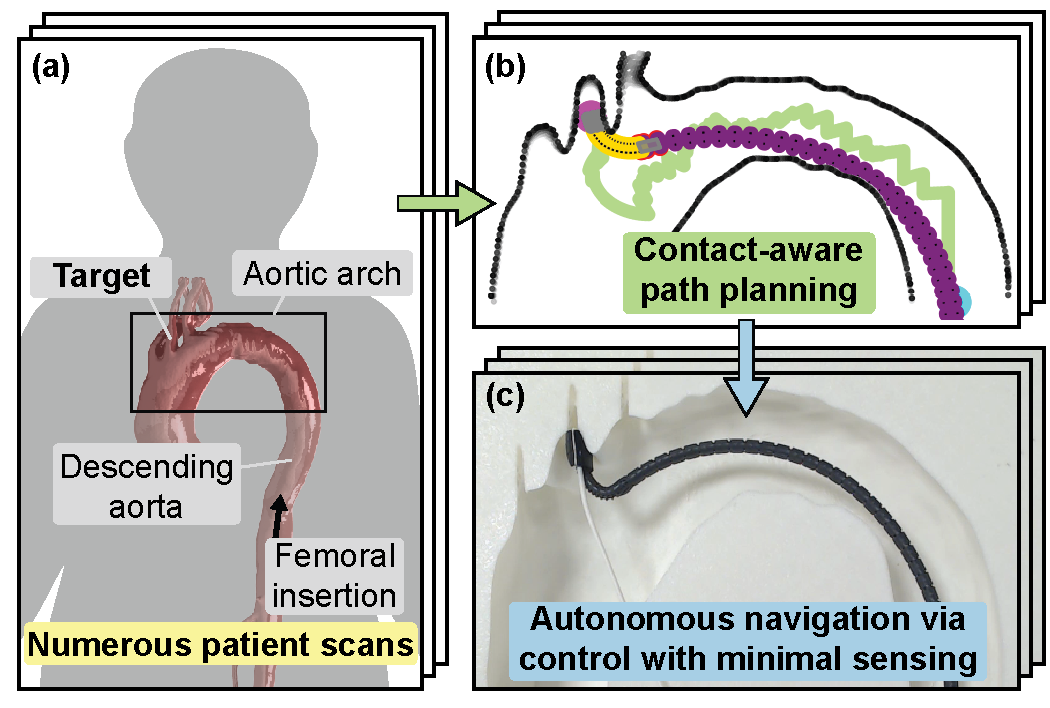}}
\vspace{-8pt}
\caption{Overview of the proposed contact-aware planning and control framework. 
(a) Patient imaging data and user-defined target location serve as inputs to 
(b) the contact-aware motion planner, which penalizes hazardous interaction and permits benign contact. 
(c) The feedback controller executes the planned trajectory using closed-loop control with minimal sensing requirements. 
The approach was validated across multiple environments derived from anatomical scans, achieving successful navigation in all three aortic arch types (see Supplementary Video S1).}
\label{fig_Purpose}  
\end{figure}

Prior planning methods tend to either avoid obstacles entirely or deliberately exploit contact to facilitate navigation \cite{patil2014needle, fairchild2021efficient, meng2022rrt, hoelscher2025safe, veil2026shape, greer2020robust, long2025csdf_iros, rao2024towards, tamhankar2026contact, gao2025parallelsimulationcontactactuation, della2020model}.
These works advanced continuum robot deployment in constrained settings but do not
(i) distinguish between benign and hazardous contact during planning, or 
(ii) integrate contact-aware, closed-loop control for reliable hardware execution. If the planned trajectory induces singularities or unstable contact modes, even sophisticated feedback control cannot guarantee successful navigation. Control strategies for continuum robots range from model-based that leverage kinematic or dynamic models to model-free (or learning based) that adapt the control using sensor feedback \cite{bajo2011configuration, goldman2014compliant, yip2014model, watson2021closed}. These controllers require kinematically feasible, contact-aware trajectories with favorable contact configurations. 

Unfavorable contact may be application specific (e.g., vascular injury during a surgery or damage to a camera during inspection).
As a representative use case, we focus on autonomous navigation of the aortic arch during mechanical thrombectomy, an endovascular procedure used to remove blood clots from cerebral vessels during an ischemic stroke. 
When clinicians treat ischemic stroke via mechanical thrombectomy, they rely on controlled contact between the body of the catheter and the environment, while deliberately avoiding contact at the tip due to the risk of vessel perforation or dissection. 
Mechanical thrombectomy consists of three primary stages: (1) navigation around the aortic arch into the target vessel, (2) traversal of the selected vasculature using a guidewire (or soft growing robot as proposed in~\cite{mangan2025serially}), and (3) clot removal from the brain.
This work focuses on stage (1), as automating navigation around the aortic arch has significant potential to expand access to care and reduce procedure time in anatomically difficult cases, thereby minimizing neurological damage from prolonged ischemia~\cite{knox2020impact, spiotta2014golden}.
To enable robust navigation in environments where contact is necessary, we introduce a unified, contact-aware planning and control approach for continuum robots that regulates contact between the robot and environment. Our contributions are summarized as follows.
\begin{itemize}
    \item \textbf{Contact-Aware Planner:} We provide a search-based planning algorithm for continuum robots that evaluates contact quality during planning, penalizing hazardous interactions while permitting benign contact.
    The planner incorporates (1) contact-aware robot transitions that ensure kinematic feasibility of the trajectory, and (2) anatomy-specific transition costs and a heuristic function that guide the selection of actions that
    mitigate contact-induced singularities and preserve robot manipulability during navigation.
    To enable realistic planning times, we introduce a task-space partitioning strategy that informs contact costs, enables a dynamic action-space, and informs a task-space based heuristic to guide the search near the goal.
    \item \textbf{Control:} We propose a trajectory tracking controller that requires only tip pose feedback.
    The controller leverages three things from the contact-aware planner: (1) precomputed tip pose waypoints, (2) joint input commands,
    and (3) contact-aware Jacobians
    to enable accurate execution with minimal sensing.
    \item \textbf{Integrated System Validation:} We demonstrate the complete framework using a multi-segment continuum robot, whose mechanical design enables navigation through diverse environments.
    We validate the system in aortic arch models created from patient scans, and, to the best of our knowledge, are the first to autonomously navigate type I, type II, and type III aortic arches while avoiding distal tip contact, thereby reducing the risk of vessel injury.
\end{itemize}

\section{Related Work}  \label{sec_related_work}

\subsection{Planning for Continuum Robots}
Motion planning for continuum robots spans both contact-agnostic and contact-aware approaches.
Contact-agnostic planners generate collision-free paths using simplified kinematic models (combined with search or sampling based planning)~\cite{patil2014needle, fairchild2021efficient, meng2022rrt, hoelscher2025safe, veil2026shape, ravigopal2021automated, abah2021image}.
These methods are computationally efficient but often assume minimal or no contact with the environment.
Contact-aware motion models better capture robot-environment interaction during planning but require more computation time.
Recent work incorporates novel heuristics, topology informed planning, and sampling based methods to reduce calls to the contact-aware motion model~\cite{rao2024towards, schegg2022automated, tamhankar2026contact, saleem2021search, wang2025topology}.
While these methods account for contact, they do not distinguish between benign and unfavorable contact, and, when deploying on real systems, do not incorporate closed-loop control to compensate for motion model errors.

\subsection{Control During Environment Interaction}
Control strategies have leveraged sensing to adapt continuum robot motion during deployment and to accommodate contact with the environment.
Model-based compliance control enables the robot to adapt its shape during insertion to comply with the environment~\cite{goldman2014compliant, abah2024self}.
Safety-constrained approaches use control barrier and Lyapunov functions~\cite{dickson2025safe, wong2025contact} to impose force limits during environment interaction.
Learning-based methods~\cite{yip2014model, watson2021closed, naughton2021elastica, chen2025data} adapt well to contact rich settings through sensing-based estimation or simulation based training.
While these strategies regulate interaction and enable adaptation to errors, they generally do not assess kinematic feasibility or differentiate trajectories based on controllability during environment interaction.

\section{Problem Statement}  \label{sec_problem_statement}

\begin{figure}
\centerline{\includegraphics[width=1.0\linewidth]{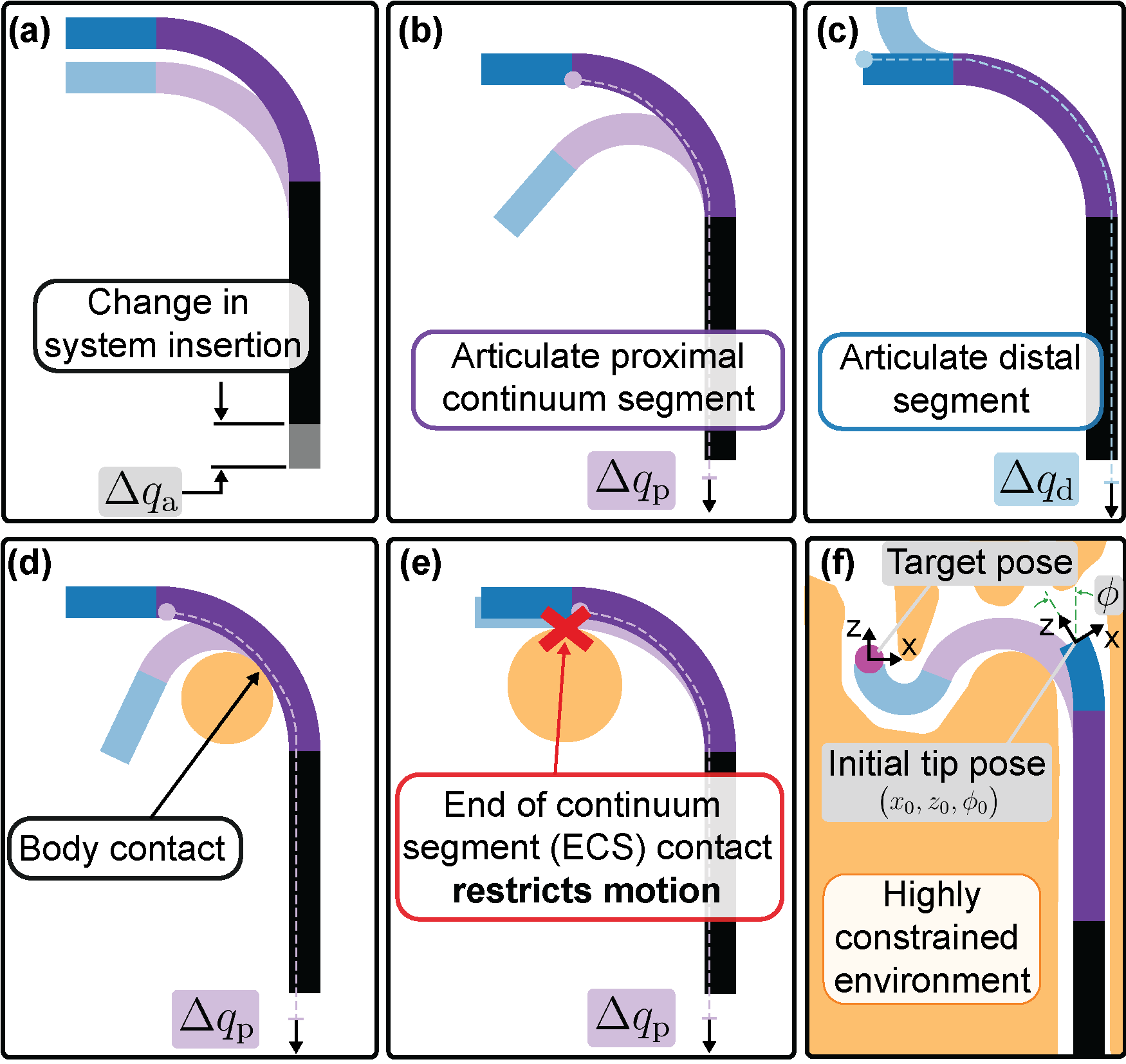}}
\vspace{-8pt}
\caption{Three joint space values $\mathbf{q} = \big[ q_{\mathrm{a}},\, q_{\mathrm{p}},\, q_{\mathrm{d}} \big]^{\top} \in \mathbb{R}^3$ control the robot (a) where $q_{\mathrm{a}}$ is axial insertion and (b-c) $q_{\mathrm{p}}, q_{\mathrm{d}}$ are the tendon displacements of the proximal and distal segments, respectively.
(d) Contact along the body of the continuum robot is acceptable while (e) contact at the end of the continuum robot (ECS) restricts motion.
(f) The objective is to navigate the continuum robot tip to a goal pose while in a highly constrained environment. 
}
\label{fig_actuator_commands}  
\end{figure}

\begin{table}[!b]
  \vspace{-8pt}
  \caption{Nomenclature}
  \label{table_notch_params}
  \centering
  \resizebox{\linewidth}{!}{%
  \begin{tabular}{c p{2.6cm} cc}
    \toprule
    Symbol & \multicolumn{3}{l}{Definition} \\
    \midrule
    \multicolumn{4}{l}{\textbf{Problem Formulation}} \\
    $\Omega$ & \multicolumn{3}{l}{Environment point cloud ($x$-$z$ projection).} \\
    $\mathcal{G} \subset  SE(2)$ & \multicolumn{3}{l}{Goal set for the robot tip pose.} \\  
    $\mathbf{q}_k \in \mathbb{R}^{3}$ 
    & \multicolumn{3}{l}{Robot joint configuration at planning time step $k$:} \\
    & \multicolumn{3}{l}{$\mathbf{q}_k = \big[ q_{\mathrm{a},k},\, q_{\mathrm{p},k},\, q_{\mathrm{d},k} \big]^{\top}$, where $q_{\mathrm{a},k}$ is axial} \\
    & \multicolumn{3}{l}{insertion, $q_{\mathrm{p},k},\, q_{\mathrm{d},k}$ are proximal/distal tendon lengths.} \\
    $\mathbf{u}_k$ & \multicolumn{3}{l}{Control input (Eq.~\eqref{eq_joint_space_update}) for change in insertion (mm),} \\
    & \multicolumn{3}{l}{and proximal/distal tendon length displacement (mm).} \\ 
    $\mathcal{F}$ & \multicolumn{3}{l}{Contact-aware motion model (Section~\ref{sec_contact_motion_model}).} \\
    $\mathcal{R}\big(q_{\mathrm{a}}, \boldsymbol{\kappa}\big) \subset \mathbb{R}^2$ & \multicolumn{3}{l}{Set of discrete points representing the robot shape with} \\
    & \multicolumn{3}{l}{$\mathcal{R}_{\mathrm{d}}$, $\mathcal{R}_{\mathrm{p}}$ representing points at the tip of the distal and} \\
    & \multicolumn{3}{l}{proximal continuum segments, respectively.} \\
    $\boldsymbol{\kappa} \in \mathbb{R}^{m_p +m_d}$ & \multicolumn{3}{l}{Vector of notch curvatures dependent on Eq.~\eqref{eq_motionmodel_opt}.} \\
    $\mathbf{p}_k \in SE(2)$ & \multicolumn{3}{l}{Robot tip pose $(x, z, \phi_{\text{tip}})$ at time step $k$}  \\
    $\tau$ & \multicolumn{3}{l}{Planning horizon to reach goal $\mathbf{p}_\tau \in \mathcal{G}$}  \\
    $\dist{A}{B}$ & \multicolumn{3}{l}{Euclidean distance between sets $A,B \subset \mathbb{R}^n$ (Eq.~\eqref{eq_dist_function})}  \\
    $\mathbf{p}^{*}_t$ & \multicolumn{3}{l}{Measured tip pose at time $t$ during online execution}  \\
    \\[-2pt]
    \multicolumn{2}{l}{\textbf{Motion Model}} & \textbf{Proximal} $(p)$  & \textbf{Distal} $(d)$ \\
    $m_p$ and $m_d$ & Number of notches     & 27       & 15 \\
    $l_i$ & Indiv. notch length          & 1.7 mm   & 1.43 mm \\
    $c$ & Uncut height          & 2.0 mm   & 1.07 mm \\
    $l_{\mathrm{s}}$ & Spacer length for tendon termination & 5.0 mm   & 5.0 mm \\
    $r$ & Radius of notched tube & 3.0 mm   & 3.0 mm \\
    \{$\mathrm{t}$\} & \multicolumn{3}{l}{Tip frame, location and orientation of $\mathbf{p}$}  \\
    \\[-8pt]
    $\boldsymbol{\xi} \in \mathcal{R}\big(q_{\mathrm{a}}, \boldsymbol{\kappa}\big)$ & \multicolumn{3}{l}{Point on robot body,
    (e.g., $\boldsymbol{\xi}_{0_{dc}}$ is point at $\{0_{dc}\}$).}  \\
    \\[-8pt]
    $l_{\text{ten}}$ & \multicolumn{3}{l}{Calculated tendon length from Eq.~\eqref{eq_prox_tendon_length}}  \\    
    \\[-2pt]
    \multicolumn{4}{l}{\textbf{Planning}} \\
    $\ell \in \mathbb{N}$ & \multicolumn{3}{l}{Number of task-space motion types required} \\
    & \multicolumn{3}{l}{to reach the goal line Eq.~\eqref{eq_action_space_partitions}.} \\
    $\mathcal{U}_{\ell}\big(\mathcal{R}\big)$ & \multicolumn{3}{l}{Partition-dependent action space (Section~\ref{sec_dynamic_action_space}).} \\ 
    $\eta_{\text{action}}(\cdot)$ & \multicolumn{3}{l}{Planning stage cost associated with joint motion.} \\
    $\eta_{\text{contact}}(\cdot)$ & \multicolumn{3}{l}{Stage cost for environment contact (Section~\ref{sec_cost_function}).} \\
    $\mathcal{T}^{(4)}$ & \multicolumn{3}{l}{The 4-connected task-space action sets.} \\
    $\mathbf{a}(x, z)$ & \multicolumn{3}{l}{Sequence of task-space motions.} \\
    & \multicolumn{3}{l}{$[\boldsymbol{a}_0, \boldsymbol{a}_1, \ldots, \boldsymbol{a}_n], \quad \boldsymbol{a}_j \in \mathcal{T}^{(4)}$} \\
    $\mathcal{P}_\ell$ & \multicolumn{3}{l}{Task-space partition region.} \\
    $\mathbf{u}_{\mathrm{step}}$ & \multicolumn{3}{l}{Discrete control increment (from hardware capabilities)} \\
    $\mathcal{H}(\boldsymbol{p}) \in \mathbb{R}$ & \multicolumn{3}{l}{Heuristic function evaluated at robot tip position $\boldsymbol{p}$.}\\
    \\[-2pt]
    \multicolumn{4}{l}{\textbf{Control}} \\
    $t$ & \multicolumn{3}{l}{Real-time index during online execution on hardware.} \\
    $\mathbf{p}^{\ast}_t$ & \multicolumn{3}{l}{Measured tip pose at time $t$.} \\
    $\mathbf{u}_{\,t}^{\text{FF}}$, $\mathbf{u}_{\,t}^{\text{FB}}$ & \multicolumn{3}{l}{Feedforward and feedback components of $\mathbf{u}_{t}$.} \\
    $\mathbf{e}_{k,t}$ & \multicolumn{3}{l}{Measured tip error $\mathbf{e}_{k,t} = \mathbf{p}_{k} - \mathbf{p}_{t}^{\ast}$} \\
    $\boldsymbol{\varepsilon} \in \mathbb{R}^2$ & \multicolumn{3}{l}{Acceptable error (position $\varepsilon_{\mathrm{pos}}$ and angle $\varepsilon_{\phi}$) during} \\
    & \multicolumn{3}{l}{trajectory tracking to move to next index $k+1$.} \\
    $T$ & \multicolumn{3}{l}{Completion time to reach the goal set $\mathbf{p}_{t}^{\ast} \in \mathcal{G}$.} \\
    \bottomrule
  \end{tabular}}
\end{table}

We consider motion planning and control of a continuum robot to autonomously navigate the aortic arch.
Table~\ref{table_notch_params} introduces the variables used in this paper.
Given a 3D anatomical scan (Fig.~\ref{fig_Purpose}(a)), which is part of the current mechanical thrombectomy workflow, and a desired goal pose for the robot tip, the objective is to reach the goal and enable deployment of a guidewire or soft growing robot~\cite{mangan2025serially}.
Orientation control is critical, as sharp approach angles can damage vessel walls during guidewire deployment~\cite{diaz2009computed}, and the engagement angle affects the performance of growing robots~\cite{haggerty2019characterizing}.
As intraoperative fluoroscopy provides only planar information, we formulate planning and control in the $x$-$z$ plane by projecting the anatomy onto this plane (black points in Fig.~\ref{fig_Purpose}(b)).
The projected anatomy is represented by $N$ points $\Omega = \left\{\boldsymbol{\omega}_n \in \mathbb{R}^2 \mid n =1, \dots, N \right\}$, and the goal is a set $\mathcal{G} \subset SE(2)$ in the 2D space of tip positions and orientations.

The continuum robot is controlled by three joint variables collected in the joint-space vector
\begin{equation}
\mathbf{q} = \big[ q_{\mathrm{a}},\, q_{\mathrm{p}},\, q_{\mathrm{d}} \big]^{\top} \in \mathbb{R}^3,
\end{equation}
where $q_{\mathrm{a}}$ is axial insertion and $q_{\mathrm{p}}, q_{\mathrm{d}}$ are the tendon lengths of the proximal and distal segments, respectively (Fig.~\ref{fig_actuator_commands}).
Each segment is antagonistically actuated by left and right tendons; for brevity we parameterize a segment by its left tendon displacement (with the right undergoing equal and opposite displacement in hardware).
At time step $k$, a control input $\mathbf{u}_k \in \mathbb{R}^3$ updates the joint configuration and produces the resulting robot curvatures $\boldsymbol{\kappa}$:
\begin{align}
\mathbf{q}_{k+1} &= \mathbf{q}_k + \mathbf{u}_k, , \label{eq_joint_space_update} \\
\boldsymbol{\kappa}_{k+1} &= \mathcal{F}\big(\mathbf{q}_{k+1}, \boldsymbol{\kappa}_{k}\big),  \label{eq_curvature_update}
\end{align}
where $\mathbf{u}_k$ represents changes in insertion length (mm), and proximal and distal continuum segment tendons displacements (mm), and 
$\mathcal{F}$ is the contact-aware motion model (Section~\ref{sec_contact_motion_model}).
The robot shape is represented by a set of discrete points $\mathcal{R}\big(q_{\mathrm{a}}, \boldsymbol{\kappa}\big) \subset \mathbb{R}^2$
which are determined by the system insertion $q_{\mathrm{a}}$ and vector of curvatures $\boldsymbol{\kappa}$.
\textbf{Planning problem:}
Given the anatomy $\Omega$, a goal set $\mathcal{G}$, and a contact-aware motion model, the objective is to compute a sequence of joint configurations $\mathbf{q}_{0:\tau}$ of arbitrary length $\tau + 1$ that drives the robot from an initial tip pose $\mathbf{p}_0 := \big(x_0, z_0, \phi_0\big)$ (where $\phi$ denotes tip angle about the $y$-axis in Fig.~\ref{fig_actuator_commands}(f)) to a goal pose $\mathbf{p}_{\tau}\in\mathcal{G}$.
The trajectory must avoid tip contact, which can cause vessel perforation, and minimize other unfavorable contact, which can induce singularities. Tip safety is enforced by requiring sufficient clearance at each planning step,
\begin{equation} \label{eq_prob_fomrulation_tipcontact}
\dist{\mathcal{R}_{\mathrm{d},k}}{\Omega} > 0, \quad \forall k \in [0,\tau],
\end{equation}
where $\mathcal{R}_{\mathrm{d},k}$ denotes the set of points at the distal tip end at step $k$ and 
\begin{equation} \label{eq_dist_function}
    \dist{A}{B} = \min_{a\in A,\; b\in B}\|a-b\|_2
\end{equation}
is the Euclidean distance between two sets. In the remainder, we also use $\dist{\mathbf{x}}{B}=\min_{b\in B}\|\mathbf{x}-b\|_2$ as the Euclidean distance between a point $\mathbf{x}$ and a set $B$.
We seek to design a planning algorithm that computes a sequence of control inputs $\mathbf{u}_{0:\tau-1}$, leading to a joint configuration trajectory $\mathbf{q}_{0:\tau}$ and associated tip pose trajectory $\mathbf{p}_{0:\tau}$ reaching the goal region $\mathcal{G}$, while satisfying the safety constraint in \eqref{eq_prob_fomrulation_tipcontact}.

\textbf{Control problem:}
Open-loop execution of the planned robot configuration sequence $\mathbf{q}_{0:\tau}$ can achieve acceptable accuracy in simple environments. However, errors in the continuum robot model and uncertainty in the environment cause deviations from the planned trajectory that propagate with insertion length and time \cite{dupont2022continuum, russo2023continuum, goldman2014compliant, oliver2021concentric, yip2014model, watson2021closed, mangan2025serially}. 
Tasks involving high curvature, long insertion distances, or complex obstacles therefore require real-time feedback control.
Given the measured tip pose $\mathbf{p}^{*}_t$ at time $t$ during online execution, the control problem is to design a real-time feedback policy that computes inputs $\mathbf{u}_t$ to drive the tip along the planned tip pose trajectory $\mathbf{p}_{0:\tau}$. 

\section{Continuum Robot Design and Contact-Aware Motion Model}  \label{sec_contact_motion_model}

This section presents the robot architecture, the piecewise constant curvature representation used to model the robot body, and a contact-aware motion model used for planning. 
The hardware design, fabrication, and actuation of the robot are presented in Section~\ref{sec_robot_design}.

\subsection{Robot Geometry}\label{sec_geometry}
To enable navigation through the aortic arch while avoiding unfavorable contact, we propose a robot with two distinct curvatures as opposed to one~\cite{rao2024towards}. The robot consists of proximal and distal tendon-driven continuum segments (purple and blue in Fig.~\ref{fig_actuator_commands}) mounted onto a linear stage for system insertion.
The continuous backbone enables smooth bending in free space and controlled compliant interaction with the surrounding anatomy. 


\begin{figure}[t]
\centerline{\includegraphics[width=1.0\linewidth]{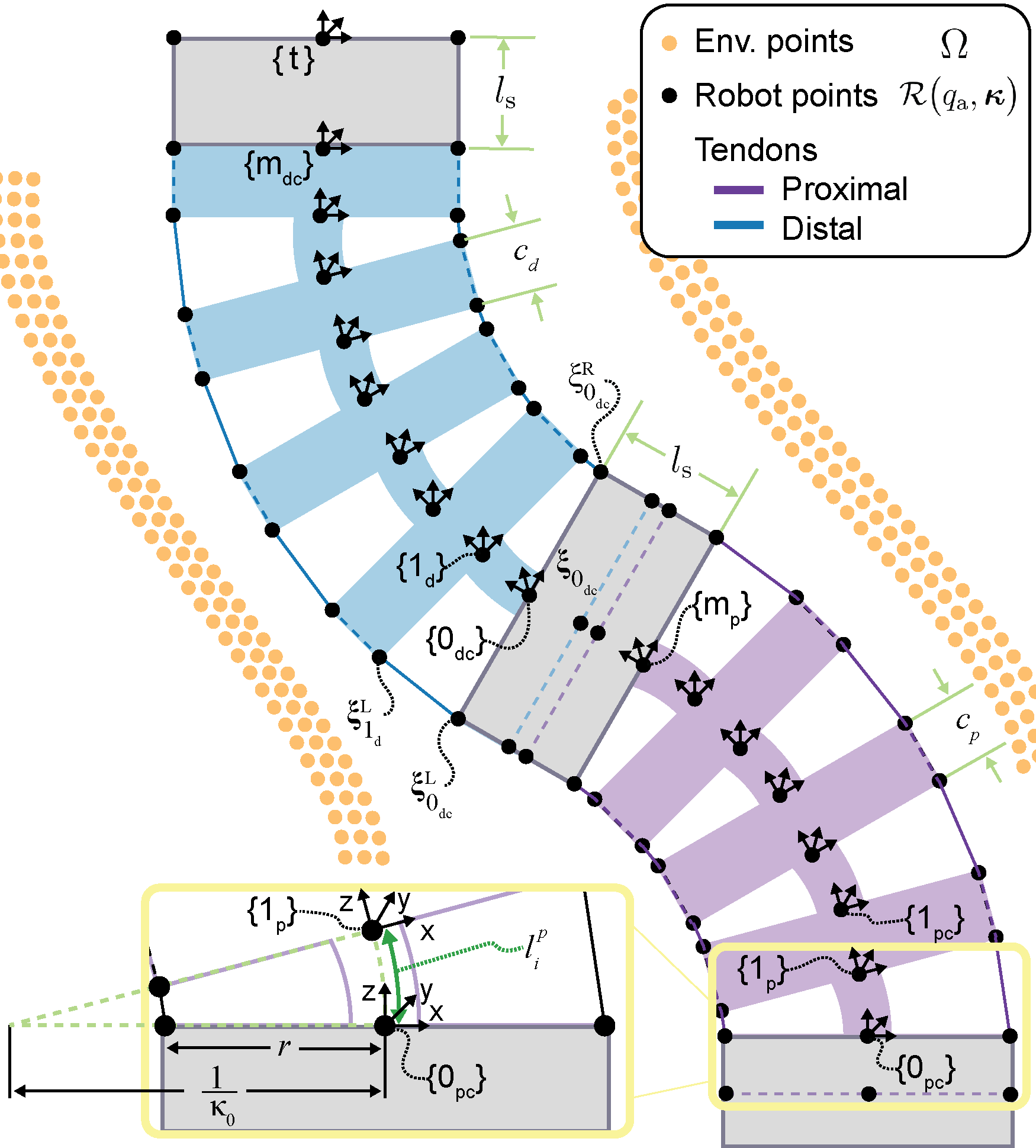}}
\vspace{-8pt}
\caption{The transformations of the piecewise constant curvature model with $m_p$ proximal notches (purple) and $m_d$ distal notches (blue).
Environment points $\Omega$ are shown in orange while robot points $\mathcal{R}\big(q_{\mathrm{a}}, \boldsymbol{\kappa}\big)$ are shown in black.
The robot geometry representation provides a structured point set to enforce contact boundaries while ensuring commanded actuation occurs through the optimization based motion model (Eq.~\eqref{eq_motionmodel_opt}).
}
\label{fig_motion_model_formulation}  
\end{figure}

We use a set of points along the body of the robot $\mathcal{R}\big(q_{\mathrm{a}}, \boldsymbol{\kappa}\big)$ to model its geometry.
Following a piecewise constant curvature assumption~\cite{york2015wrist}, the homogeneous transformation of the $i$-th notch is denoted by $T_{i_c}^{i+1}$, where $i_c$ is the frame at the base of the $i$-th notch, located above the uncut height $c$, and $i+1$ is the frame at the distal end of that notch.
Figure~\ref{fig_motion_model_formulation} illustrates this for the transformation from frame $\{0_{pc}\}$ to $\{1_{p}\}$ where $p$ indicates the proximal continuum segment.
With curvature $\kappa_i$ and arc length $l_i$ the homogeneous transformation is 
\begin{equation} \label{eq_constcurv_Trans}
T_{i_c}^{i+1}(\kappa_i) =
\begin{bmatrix}
R_y(\theta_i) & p_i \\
0 & 1
\end{bmatrix},
\quad
p_i =
\begin{bmatrix}
\dfrac{1 - \cos\theta_i}{\kappa_i} \\
\dfrac{\sin\theta_i}{\kappa_i}
\end{bmatrix}.
\end{equation}
where $\theta_i = \kappa_i l_i$ and $R_y(\theta_i) \subset SO(2)$ denotes rotation about the $y$-axis.
Let $T_{z,c}$ denote pure translation along $z$ by uncut height $c$, with subscripts $p$ and $d$ representing the proximal and distal segments respectively, and $l_{\mathrm{s}}$ be the length of the spacer for tendon terminations (Fig.~\ref{fig_motion_model_formulation}).
Composing these transformations, the transformation from the base of proximal continuum segment \{$\text0_{pc}$\} to its end \{$m_{p}$\}, and from \{$m_{p}$\} to the robot tip frame \{$\mathrm{t}$\}, are:
\begin{gather} 
 T_{0_{pc}m_{p}}
=
  \left(\prod_{i=0}^{m_p - 1} T_{i_c}^{i+1} T_{z,c_p}\right)T_{z,-c_p}, \label{eq_proximal_seg}\\
 T_{m_{p}{\mathrm{t}}}
 =
  T_{z,l_{\mathrm{s}}}\left(\prod_{i=m_p}^{m_p + m_d - 1} T_{i_c}^{i+1} T_{z,c_d}\right)T_{z,-c_d}T_{z,l_{\mathrm{s}}}, \label{eq_distal_seg}
\end{gather}
where $m_p$ and $m_d$ are the notch counts of the proximal and distal segments, respectively. 
All relevant design variables are summarized in Table~\ref{table_notch_params}.

The shape of the robot $\mathcal{R}\big(q_{\mathrm{a}}, \boldsymbol{\kappa}\big)$ is determined by the static parameters above (e.g., $l_i$, $c$, $l_\mathrm{s}$), the individual curvatures $\boldsymbol{\kappa} \in \mathbb{R}^{m_p +m_d}$, and the axial insertion of the robot $q_{\mathrm{a}}$.
As shown in Fig.~\ref{fig_motion_model_formulation}, the robot is represented by a set of back bone points derived from ~\eqref{eq_proximal_seg} and~\eqref{eq_distal_seg}, together with the edge points obtained by translating each backbone point laterally by the robot radius $r$ in both directions (e.g., $\boldsymbol{\xi}_{0_{dc}}^L$ and $\boldsymbol{\xi}_{0_{dc}}^R$ for frame $\{0_{dc}\}$ are shown in Fig.~\ref{fig_motion_model_formulation}).
The contact-aware motion model (Section~\ref{sec_opt_based_motion_model}) incorporates these points in the optimization constraints to model contact with the environment and enforce consistency between the commanded and realized tendon lengths.

\subsection{Contact-Aware Motion Model} \label{sec_opt_based_motion_model}
We adapt a contact-aware motion model in~\cite{ashwin2021profile,rao2024towards} to a two-segment notched continuum robot and extend it to handle non-convex environment geometry.
The model maps commanded joint values to robot curvatures while enforcing contact and actuation consistency.
Given the joint-space commands, $\mathbf{q}_k = \big[ q_{\mathrm{a},k},\, q_{\mathrm{p},k},\, q_{\mathrm{d},k} \big]^{\top}$, the contact-aware motion model $\mathcal{F}(\cdot)$ computes the updated curvatures (Eq.~\eqref{eq_curvature_update}).
The contact-aware motion model solves an energy-minimization problem while enforcing environmental and actuation constraints:
\begin{argmini!}[3] 
{\boldsymbol{\kappa}}{\sum_{i=0}^{m_p+m_d-1} (\kappa_i)^2}
{\label{eq_motionmodel_opt}}
{\mathcal{F}\big(\mathbf{q}_{k}, \boldsymbol{\kappa}_{k-1}\big)
=}
\addConstraint{\dist{\mathcal{R}(q_{\mathrm{a},k},\boldsymbol{\kappa})}{\Omega}}{\geq \gamma_1}\label{eq_motionmodel_opt_constraint_env}
\addConstraint{l_{\text{ten}}^p\big(\mathcal{R}(q_{\mathrm{a},k},\boldsymbol{\kappa})\big) - q_{\mathrm{p},k}}{= 0}\label{eq_motionmodel_opt_constraint_qprox}
\addConstraint{l_{\text{ten}}^d\big(\mathcal{R}(q_{\mathrm{a},k},\boldsymbol{\kappa})\big) - q_{\mathrm{d},k}}{= 0}\label{eq_motionmodel_opt_constraint_qdist}
\end{argmini!}

The first constraint Eq.~\eqref{eq_motionmodel_opt_constraint_env} ensures that the Euclidean distance between the points on the robot $\mathcal{R}(q_{\mathrm{a},k},\boldsymbol{\kappa})$ and the points representing the environment $\Omega$ (orange in Fig.~\ref{fig_motion_model_formulation}) remains above some small threshold $\gamma_1$.
The second and third constraints  Eq.~\eqref{eq_motionmodel_opt_constraint_qprox}-\eqref{eq_motionmodel_opt_constraint_qdist} enforce consistency between the commanded tendon lengths ($q_{\mathrm{p},k}$ and $q_{\mathrm{d},k}$) and the calculated tendon lengths computed from the resulting robot shape, ensuring that the optimized configuration reflects the applied joint commands.
For the uncut heights ($c_p$ and $c_d$ in Fig.~\ref{fig_motion_model_formulation}) the tendon is mechanically constrained, so the tendon length is equal to the respective uncut height. 
For the notched sections, the tendon is not mechanically constrained and thus the Euclidean distance is found between the robot points. The calculated tendon length for the proximal continuum segment is
\begin{equation} \label{eq_prox_tendon_length}
    l_{\text{ten}}^p(\mathcal{R}(q_{\mathrm{a}},\boldsymbol{\kappa})) =
    c_p(m_p - 1) +
    \sum_{i=0}^{m_p - 1} \|\boldsymbol{\xi}_{i_{pc}}^L - \boldsymbol{\xi}_{i+1_{p}}^L\|_2 .
\end{equation}
The calculated tendon length of the distal continuum segment is found through the same formula but with $d$ replacing the $p$ sub- and super-scripts.
Because there are multiple potential solutions to the constrained optimization, we use the previous curvature values $\boldsymbol{\kappa}_{k-1}$ for the seed when solving the optimization in~Eq.~\eqref{eq_motionmodel_opt}. 
Section~\ref{sec_contact_planner} describes how this model is incorporated into search-based planning during node expansion.

\section{Contact-Aware Planning} \label{sec_contact_planner}
\begin{figure}[t]
\centerline{\includegraphics[width=1.0\linewidth]{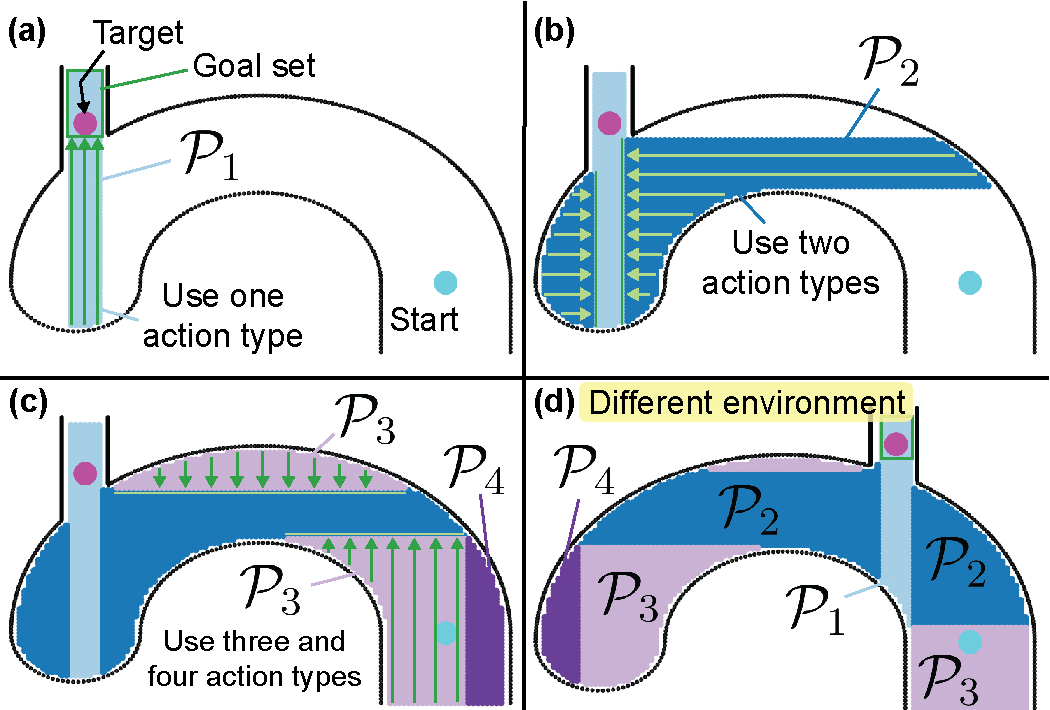}}
\vspace{-8pt}
\caption{Illustration of the task-space partitions and their relationship to the target vessel. 
(a) The green box marks the goal set ($\mathcal{G}$). 
(b-c) Each partition groups cells that can reach the goal set location using the same minimal sequence of task-space action types (e.g., $\mathcal{P}_1$ requires one task-space action type). 
(d) Task-space partitions for another environment.
The task-space partitions let the contact-aware continuum robot motion planner select a dynamic action set $\mathcal{U}_\ell$ appropriate to the local task-space complexity, and inform the contact cost function.}
\label{fig_task_space_partitions}  
\end{figure}

The objective of the planning module is to compute a contact-aware, kinematically feasible plan for a continuum robot navigating a highly constrained environment.
We formulate this as a search-based motion planning problem on the robot's configuration space discretized as a graph $G = (\mathcal{V}, E)$:
\begin{align}
\min_{v_{0:\tau},\tau} \quad &
\sum_{k=0}^{\tau-1} 
\eta_{\text{action}}(\mathbf{q}_{v_k},\mathbf{q}_{v_{k+1}})
+ \eta_{\text{contact}}\big(\mathcal{R}(q_{\mathrm{a},v_{k+1}}, \boldsymbol{\kappa}_{v_{k+1}})\big) \notag\\
\text{s.t.} \quad 
& \mathbf{p}(q_{\mathrm{a},v_\tau},\boldsymbol{\kappa}_{v_\tau}) \in \mathcal{G},\notag\\
& (v_k, v_{k+1}) \in \mathcal{U}_{\ell}\big(\mathcal{R}(q_{\mathrm{a},v_{k}},\boldsymbol{\kappa}_{v_{k}})\big) \subseteq E,\notag\\
& \boldsymbol{\kappa}_{v_{k+1}} = \mathcal{F}\big(\mathbf{q}_{v_{k+1}}, \boldsymbol{\kappa}_{v_k}\big). \label{eq_planning_formulation}
\end{align}
Here, $\mathbf{q}_{v_k}$ and $\boldsymbol{\kappa}_{v_k}$ denote the joint configuration and curvature associated with each graph node $v_k \in \mathcal{V}$. The first constraint requires that the distal tip pose $\mathbf{p}$ reaches the goal set $\mathcal{G}$ at the last configuration $\mathbf{q}_{v_\tau}$.
In the second constraint, we impose a partition-dependent action space $\mathcal{U}_{\ell}$ in addition to the edge set $E$ (Section~\ref{sec_dynamic_action_space}). 
The third constraint propagates the curvatures $\boldsymbol{\kappa}_{v_{k+1}}$ with the contact-aware motion model $\mathcal{F}(\cdot)$ (Eq.~\eqref{eq_motionmodel_opt}). 
The objective penalizes both joint motion $\eta_{\text{action}}(\mathbf{q}_{v_k},\mathbf{q}_{v_{k+1}})$ and environment interaction evaluated from the resulting robot geometry $\eta_{\text{contact}}(\mathcal{R}_{v_{k+1}})$ (Section~\ref{sec_cost_function}).
The resulting optimal sequence of graph nodes $v_{0:\tau}^\star$ (i.e., the solution of Eq.~\eqref{eq_planning_formulation}, Alg~\ref{alg_planning}, line~\ref{alg:return_optimal}) defines the reference trajectories for joint configuration, feedforward control action, and tip pose, respectively as $\mathbf{q}_{k} = \mathbf{q}_{v_k^\star}$,  $\mathbf{u}_k = \mathbf{q}_{k+1} - \mathbf{q}_{k}$ and $\mathbf{p}_{k} = \mathbf{p}_{v_k^\star}$, which are tracked by the controller in Section~\ref{sec_control}.

We first describe the discretization and the node representation.
Next, we introduce task-space partitions that structure the workspace, enabling a dynamic, partition dependent action space that improves planning efficiency while differentiating regions where contact is acceptable and informing contact cost.
We then detail the action and contact cost functions and their physical interpretation.
Finally, we present a task-space heuristic which guides the robot tip and computational acceleration strategies within the search-based planning framework.
Algorithm~\ref{alg_planning}
summarizes the overall contact-aware planning procedure using a weighted A* search guided by precomputed task-space partitions and a heuristic field.
Supplementary Video S1 also shows the planning process.

\begin{algorithm}[t]
\caption{Contact-Aware Planning (Weighted A*)}
\label{alg_planning}
\begin{algorithmic}[1]
\Require anatomy $\Omega$, initial configuration and curvature $\mathbf{q}_{0}$, $\boldsymbol{\kappa}_0$, goal set $\mathcal{G}$, weight $\epsilon \ge 1$
\State $\{\mathcal{P}_\ell\} \leftarrow$ PartitionWorkspace$(\Omega, \mathcal{G})$ \Comment{Section~\ref{sec_task_space_partitions}}
\State $\mathcal{H} \leftarrow$ ComputeHeuristic$(\Omega, \mathcal{G}, \{\mathcal{P}_\ell\})$ \Comment{Section~\ref{sec_heuristic}}

\State OPEN $\leftarrow$ priority queue keyed by $f = g + \epsilon \mathcal{H}$
\State Create start node $v_0 \gets (\mathbf{q}_{0}, \boldsymbol{\kappa}_{0})$,  $g[v_0] \leftarrow 0$
\State push $(v_0, f = \epsilon \, \mathcal{H}($tipPos$(\mathcal{R}(q_{\mathrm{a},v_0},\boldsymbol{\kappa}_{v_0})$$)$ into OPEN

\While{OPEN not empty}
    \State $v \leftarrow$ pop lowest-$f$ node from OPEN
    \If{ tipPose$(\mathcal{R}(q_{\mathrm{a},v},\boldsymbol{\kappa}_{v})\in \mathcal{G}$} 
    \State \Return $v_{0:\tau}^\star =$ reconstructPath$(v)$ \EndIf\label{alg:return_optimal}

    \State $\mathcal{P}_{\ell,v} \leftarrow$ PartitionLookup$(\mathbf{p}_v)$
    \ForAll{$v'$ \textbf{in} getSuccessors$(v, \mathcal{P}_{\ell,v})$} 
        \State solve $\boldsymbol{\kappa}' \leftarrow \mathcal{F}(\mathbf{q}_{v'}, \boldsymbol{\kappa}_v)$ \label{alg:motion_model}
        \State \Comment{Contact-aware motion model (Section~\ref{sec_contact_motion_model})}
        \State Store $v' \gets (\mathbf{q}_{v'}, \boldsymbol{\kappa}_{v'})$. 

        \State $c \leftarrow \eta_{\text{action}}(\mathbf{q}_v,\mathbf{q}_{v'}) + \eta_{\text{contact}}(\mathcal{R}(q_{\mathrm{a},v'},\boldsymbol{\kappa}_{v'}))$
        \If{$g[v']$ undefined \textbf{or} $g[v'] > g[v] + c$}
            \State $g[v'] \leftarrow g[v] + c$
            \State $f \leftarrow g[v'] + \epsilon \, \mathcal{H}($tipPos$(\mathcal{R}(q_{\mathrm{a},v'},\boldsymbol{\kappa}_{v'})$$)$
            \State update $(v', f)$ in OPEN
        \EndIf
    \EndFor
\EndWhile
\State \Return \textbf{failure}
\end{algorithmic}
\end{algorithm}

\subsection{Discretization and Node Representation}
To apply search-based methods to motion planning of a continuum robot, we discretize the robot's configuration space $\mathcal{Q}$ into a graph $G = (\mathcal{V}, E)$.
The nodes $\mathcal{V}$ correspond to a regular 3D grid over the configuration space, with bounds $\mathbf{q}_{\text{min}}$ and $\mathbf{q}_{\text{max}}$, and resolution $\mathbf{u}_{\mathrm{step}} \in \mathbb{R}^{3}$ in each dimension. These represent the insertion/length actuation bounds and increments available in hardware.
The edge set $E$ initially connects all adjacent neighbors within a $3\times3\times3$ window, consisting of 26 neighbors per node overall. For efficiency, the edge set $E$ is further restricted by a partition-dependent action set $\mathcal{U}_{\ell}$ described later.

An important distinction from other search-based methods is that we associate each node $v \in \mathcal{V}$ with a unique curvature vector $\boldsymbol{\kappa}_{v}$, propagated using the contact-aware motion model $\mathcal{F}(\cdot)$ (Eq.~\eqref{eq_motionmodel_opt}) from the initial curvature $\boldsymbol{\kappa}_0$ at the initial configuration $\mathbf{q}_{0}$.
Although identical actuator values may correspond to multiple physical robot shapes~\cite{rao2024towards}, 
prior work has shown that configuration spaces can be decomposed into regions associated with distinct local optima during trajectory optimization~\cite{saleem2021search, wang2025topology}. These regions closely relate to topologically distinct path families that cannot be continuously deformed into one another~\cite{bhattacharya2012topological}. 
In the environments considered here, $\Omega$ contains no floating obstacles and the robot backbone forms a continuous curve.
Under this assumption, the mapping between joint variables and robot deformation is treated as locally unique, allowing each node to be parameterized solely by the joint configuration $\mathbf{q}$.
Compared to~\cite{rao2024towards}, this reduces dimensionality by $(m_p + m_d - 3)$ (43 dimensions for our robot) and improves planning efficiency.

\subsection{Task-Space Partitions} \label{sec_task_space_partitions}
The node representation described above assumes joint configurations uniquely determine robot shape within the region of interest.
However, even within the same workspace region, different areas may require different control effort or allowable contact interactions. 
For instance, at the beginning of a plan, the continuum robot may need to primarily insert and articulate one continuum segment while avoiding tip contact, while closer to the goal, more joint inputs need to be considered and tip contact may be necessary to achieve the goal (in our mechanical thrombectomy case of entering the origin of the internal carotid artery).
To capture these differences, we partition the task space according to the minimal number of task-space motion types required to reach the goal (Fig.~\ref{fig_task_space_partitions}).

Unlike previous planners that primarily distinguish regions based on geometric distance or collision feasibility, the proposed partitions characterize workspace regions by their required motion complexity. 
This enables the planner to distinguish between regions that are geometrically connected and serves three primary purposes: 
(1) determining the dynamic action space ($\mathcal{U}_{\ell}$ in Eq.~\eqref{eq_planning_formulation}),
(2) shaping contact stage costs (Section~\ref{sec_cost_function}), and
(3) guiding a safer approach direction near bifurcations (Section~\ref{sec_heuristic}).


\subsubsection{Partition Construction}
To construct the partitions, we discretize the 2D projected workspace and assign each cell according to the minimal task-space motion complexity required to reach the position of the goal set $\mathcal{G}_{\mathrm{pos}}$. 
Each cell $\boldsymbol{p} = [x, z]^\top$ represents a task space location.
Feasible paths respect collision constraints and move within a 4-connected action grid:
\[
\mathcal{T}^{(4)} = \{ (\pm1,0),\ (0,\pm1) \} 
= \{\textit{up},\ \textit{down},\ \textit{left},\ \textit{right}\}.
\]
From every cell $\boldsymbol{p}$, we identify feasible paths to the goal $\mathcal{G}_{\mathrm{pos}}$ satisfying these constraints. 
As illustrated in Fig.~\ref{fig_task_space_partitions}, the $\mathcal{G}_{\mathrm{pos}}$ is the green box in the target vessel, and each partition groups cells requiring the same minimal number of motion-type switches.
Cells reaching the goal with a single motion type (e.g., only upward motion) belong to $\mathcal{P}_1$.
Cells requiring one switch (e.g., left $\rightarrow$ up) belong to $\mathcal{P}_2$, and higher index partitions correspond to increasingly complex motion sequences.
This construction partitions the workspace by task-space control complexity rather than geometric distance, allowing the planner to reason about the minimum number of tip directional changes needed to reach the goal.

\subsubsection*{Formal Definition}
We denote the sequence of task-space actions that form a feasible path from an initial point $\boldsymbol{p}_{0}~=~(x,z)$ to a goal set position $\mathcal{G}_{\mathrm{pos}}$ as 
\begin{equation}
    \mathbf{a}(x, z) = [\boldsymbol{a}_0, \boldsymbol{a}_1, \ldots, \boldsymbol{a}_n], \quad \boldsymbol{a}_j \in \mathcal{T}^{(4)}.
\end{equation}
The minimum number of motion types required to reach the goal is: 
\begin{equation} \label{eq_action_space_partitions}
\begin{aligned}
\ell(x, z) = \min_{\mathbf{a}(x, z)} \quad & 1 + \sum_{j=0}^{n-1} \mathbbm{1}\{\boldsymbol{a}_{j} \neq \boldsymbol{a}_{j+1}\} \\
\text{subject to} \quad 
& \boldsymbol{p}_{j+1} = \boldsymbol{p}_{j} + \boldsymbol{a}_{j}, \\
&\dist{\boldsymbol{p}_{j}}{\Omega} > 0, \quad \forall k \in [0,n] \\
& \boldsymbol{p}_{n} \in \mathcal{G}_{\mathrm{pos}},
\end{aligned}
\end{equation}
where the second constraint specifies that all points in the path must be collision free.
The task-space partitions $\mathcal{P}_\ell$ for all cells in the workspace are then defined based on Eq.~\eqref{eq_action_space_partitions}.

\subsubsection{Partition-Dependent Action Spaces} \label{sec_dynamic_action_space}
For planning, we assign each partition $\mathcal{P}_\ell$ a corresponding discrete action set $\mathcal{U}_\ell$, which restricts the available motion primitives in the robot’s actuation space. 
During planning, the action set is chosen as:
\begin{equation}
    \mathbf{p} \in \mathcal{P}_\ell \ \Rightarrow\  \text{use action set } \mathcal{U}_\ell.
\end{equation}
This mapping lets the planner adapt the motion primitives to the local task-space complexity, improving planning efficiency.
The choice of action set $\mathcal{U}_\ell$ depends on the robot's geometry and construction.
In our mechanical thrombectomy case, for example, while in the descending aorta, we allow only insertion and actuation of the distal continuum segment in $\mathcal{P}_{4}$, and this action space expands when close to the target vessel in $\mathcal{P}_{1}$ (Fig.~\ref{fig_task_space_partitions}).

\subsection{Cost Function} \label{sec_cost_function}
In Eq.~\eqref{eq_planning_formulation}, $\eta_{\text{action}}$ denotes the cost of executing a joint command.
Action costs are proportional to the corresponding control increments and their respective affect on tip motion:
\begin{equation}
\eta_{\text{action}} =
\begin{cases}
c_{\mathrm{a}}\cdot\,\mathrm{u}_{\mathrm{step, a}} & \text{insertion}, \\
c_{\mathrm{p}}\cdot\,\mathrm{u}_{\mathrm{step, p}}, & \text{proximal segment (longest)}, \\
c_{\mathrm{d}}\cdot\,\mathrm{u}_{\mathrm{step, d}}, & \text{distal segment (shortest)}.
\end{cases}
\end{equation}
where $c_{\mathrm{a}}$, $c_{\mathrm{p}}$, $c_{\mathrm{d}}$ denote the tip displacement resulting from a unit actuation of each input in free space.
For our robot, these weights are set to $c_{\mathrm{a}} = 1.1$, $c_{\mathrm{p}} = 60$, and $c_{\mathrm{d}}=10$, based on the geometry in Table~\ref{table_notch_params} and Eq.~\eqref{eq_motionmodel_opt}.
A result of this choice is that the action costs exceed the distance of the associated free-space distal tip displacement,
\[
\eta_{\text{action}}(\mathbf{q}_k,\mathbf{q}_{k+1}) >
\lVert \mathbf{p}_{k+1} - \mathbf{p}_{k} \rVert_2 ,
\]
which makes the task-space distance heuristic (Section~\ref{sec_heuristic}) consistent~\cite{hart1968formal}.
Empirically, free-space produces larger tip displacement than motion under environmental contact, except for rare snapping behaviors (analogous to bending a finger against a surface and sliding it off), which are excluded from our search process.

The contact cost, $\eta_{\text{contact}}(\mathcal{R})$ discourages unfavorable environment interaction:
\begin{equation}\label{eq_contact_cost}
    \begin{aligned}
        \eta_{\text{contact}}(\mathcal{R}) = 
        & \ C_{\mathrm{d}}\mathbbm{1}[\dist{\mathcal{R}_{\mathrm{d}}}{\Omega}\le\gamma_{\text{ECS}}] \ + \\
        & C_{\mathrm{p}}\mathbbm{1}[\dist{\mathcal{R}_{\mathrm{p}}}{\Omega}\le\gamma_{\text{ECS}}] +
        C_{\text{body}}\frac{N_{\text{contact}}}{R_{\max}}.
    \end{aligned}
\end{equation}
Here $\mathcal{R}_{\mathrm{d}}$ and $\mathcal{R}_{\mathrm{p}}$ denote point sets at the distal tip and proximal segment end, and $N_{\text{contact}}$ counts robot points within distance $\gamma_{\text{contact}}$ of the environment. 
We choose $C_{\mathrm{d}}\!\gg\!C_{\mathrm{p}}\!\gg\!C_{\text{body}}$, prioritizing avoidance of distal tip contact due to vessel perforation risk (Eq.~\eqref{eq_prob_fomrulation_tipcontact} in problem formulation).
End of continuum segment (ECS) is the next most critical, as it constrains segment motion and reduces tip displacement achievable from actuation
(Fig.~\ref{fig_actuator_commands}(e)), degrading robot manipulability through loss of Jacobian rank~\cite{lynch2017modern}.
Body contact is permitted but  minimized to limit unnecessary interaction and potential damage. 
Because controlled tip contact is required when entering the target vessel, contact penalties are disabled in the final task-space partition $\mathcal{P}_1$ when the tip orientation is within the goal orientation range $|\phi_{\text{tip}}|\le\mathcal{G}_{\phi}$.
Since planning is search-based rather than gradient-based, this state-dependent modification does not affect optimization stability.

\subsection{Task-Space Distance Heuristic} \label{sec_heuristic}

\begin{figure}[t]
\centerline{\includegraphics[width=1.0\linewidth]{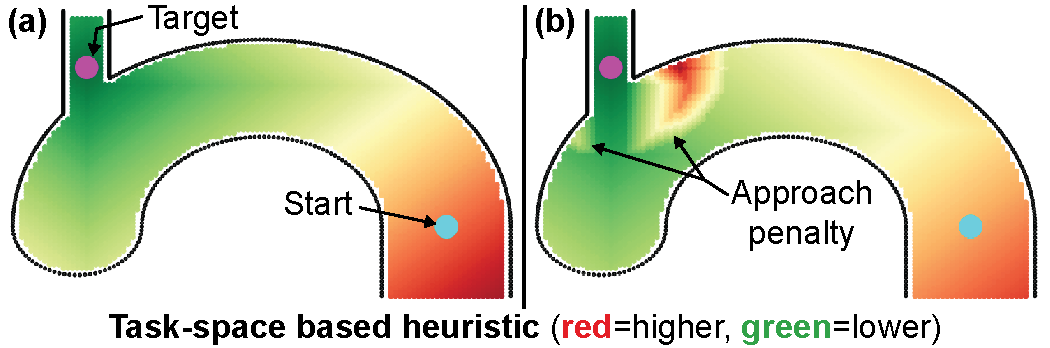}}
\vspace{-8pt}
\caption{(a) The Dijkstra planner generates a heuristic field that encodes move cost and environment proximity. 
(b) The heuristic prioritizes cells in task-space partition $\mathcal{P}_1$ (Fig.~\ref{fig_task_space_partitions}(e)).
The approach penalty reduces articulation near bifurcation points which improves robustness in hardware experiments.
}
\label{fig_heuristic}  
\end{figure}

We construct a task-space heuristic combining three elements: (i) estimated effort to reach the goal using a task-space motion model, 
(ii) collision pruning, and 
(iii) an approach penalty that promotes axial alignment of the robot tip with the target vessel at bifurcation points.
The heuristic $\mathcal{H}(\boldsymbol{p}) \in \mathbb{R}$ assigns a scalar cost to each workspace cell $\boldsymbol{p} = (x, z)$ (Fig.~\ref{fig_heuristic}). Using the same 4-connected action grid and constraints described in Eq.~\eqref{eq_action_space_partitions}, we compute
\begin{equation}\label{eq_heuristic}
\mathcal{H}(x,z) =
\min_{\mathbf{a}(x,z)}
\sum_{j=0}^{n-1}
\Big(
c_{\text{action}}(\boldsymbol{a}_j)
+
c_{\text{approach}}(\boldsymbol{p}_j)
\Big)
\end{equation}
subject to
\[
\boldsymbol{p}_{j+1} = \boldsymbol{p}_j + \boldsymbol{a}_j, \quad
\dist{\boldsymbol{p}_j}{\Omega} > 0,\quad
\boldsymbol{p}_n \in \mathcal{G}_{\mathrm{pos}}.
\]
Here $c_{\text{action}}(\boldsymbol{a}_j)$ is a Euclidean step cost, and the approach penalty
\begin{equation}
c_{\text{approach}}(\boldsymbol{p}) =
C_{\text{app}}\,
\mathbbm{1}\!\left[
\dist{\boldsymbol{p}}{\mathcal{G}_{\mathrm{pos}}} < r_{\text{approach}}
\ \& \
\boldsymbol{p}\notin\mathcal{P}_1
\right]
\end{equation}
discourages lateral approaches near the goal while encouraging entry through $\mathcal{P}_1$ (Fig.~\ref{fig_heuristic}(b)).
The heuristic efficiently guides planning through anatomically feasible paths while avoiding vessel bifurcation points.
Values are computed via a backward Dijkstra expansion from the goal in under two seconds.

\subsection{Planning Acceleration Techniques} \label{sec_planning_acc_techniques}
The contact-aware motion model optimization operates in a high dimensional space ($m_p+m_d=46$) and is evaluated thousands of times during planning, so we provide several acceleration strategies to achieve practical planning times.
First, we use analytical derivatives within the gradient-based optimizer, which substantially reduce function evaluations compared to gradient-free methods and finite-difference approximations, particularly as the number of input variables scales~\cite{hwang2015modular}. 
The full derivative formulations are provided in the Appendix (Section~\ref{sec_derivatives}).
Second, we implement function evaluation caching to reduce calls to the contact-aware motion model during planning.
Finally, we impose reduced iteration limits for the motion model optimization to cap computation time.
Joint inputs that fail to converge to curvature solutions from Eq.~\ref{eq_motionmodel_opt} within these limits are discarded and excluded from node expansion in the search algorithm (i.e., skipped after Alg.~\ref{alg_planning}, line~\ref{alg:motion_model}).
This pruning strategy prevents the planner from investing computation in infeasible or poorly conditioned configurations. 
If the limits need to expand around the goal region, the iteration limits can be dynamic and can depend on the task-space partitions $\mathcal{P}_{\ell}$, but we did not find this necessary in our use case.
\begin{figure*}[t!]
\centerline{\includegraphics[width=1.0\linewidth]{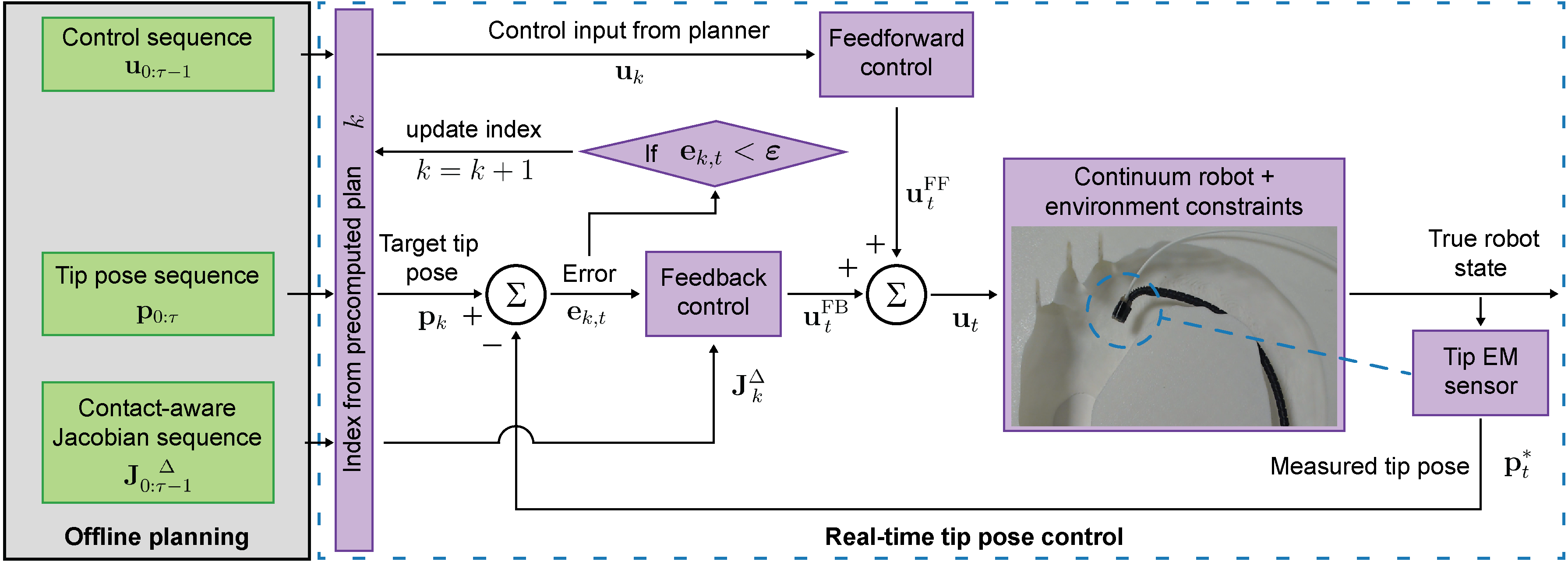}}
\vspace{-8pt}
\caption{The proposed controller tracks a trajectory of target poses target poses $\mathbf{p}_k$ using the measured tip pose $\mathbf{p}^{\ast}_t$ as feedback, where $k$ denotes the discrete planning index along the trajectory and $t$ denotes the real-time control index during hardware execution.
The contact-aware planner first operates offline using a robot and environment model, and desired goal to generate three outputs:
(1) a sequence of kinematically feasible target poses $\mathbf{p}_{0:\tau}$,
(2) control inputs $\mathbf{u}_{0:{\tau-1}}$ from the plan used for feedforward control, and
(3) a sequence of contact-aware Jacobians $\mathbf{J}^{\ \ \Delta}_{0:{\tau-1}}$ for feedback control.}
\label{fig_control_schematic}  
\end{figure*}

\section{Control} \label{sec_control}
Precomputed joint values from the contact-aware motion planner provide sufficient accuracy for simple motions in localized environments. 
However, tasks involving high curvatures or long trajectories require closed-loop control to compensate for tendon friction, inter-segment coupling, and environment misalignment~\cite{dupont2022continuum, russo2023continuum, goldman2014compliant, oliver2021concentric, yip2014model, watson2021closed, mangan2025serially}.
We use $k$ to denote the discrete planning index for the nominal trajectory, and $t$ to denote the real-time index during online execution on hardware.
Below we present the method for computing the control input $\mathbf{u}_{t}$ that steers the measured tip pose $\mathbf{p}^{\ast}_t$ along a trajectory of target poses $\mathbf{p}_k$, for $k = 0, 1, \dots, \tau$ (Fig.~\ref{fig_control_schematic}).
The trajectory tracking control combines feedforward ($\mathbf{u}_{\,t}^{\text{FF}}$) and feedback ($\mathbf{u}_{\,t}^{\text{FB}}$) components:
\begin{equation}
    \mathbf{u}_{\,t} = 
    \beta \, \mathbf{u}_{\,t}^{\text{FF}} +
    (1 - \beta) \, \mathbf{u}_{\,t}^{\text{FB}},
    \label{eq_total_control}
\end{equation}
where $\beta \leq 1$ is a static scaling factor found empirically.
We also detail the computation of a contact-aware robot Jacobian, which enables feedback control in constrained environments.

\subsection{Trajectory Tracking} \label{sec_trajectory_control}
While tracking target pose $\mathbf{p}_{k}$ at time $t$, the measured tip error is 
\begin{equation}
    \mathbf{e}_{k,t} = \mathbf{p}_{k} - \mathbf{p}_{t}^{\ast}.
    \label{eq_measured_error}
\end{equation}
When the positional error magnitude $\| (e_{x,k,t},\,e_{z,k,t}) ^{\top}\|_2$ falls below the distance threshold $\varepsilon_{\mathrm{pos}}$ and the angular error satisfies $|e_{\phi,k,t}| \le \varepsilon_{\phi}$, the controller advances to the next target $\mathbf{p}_{k+1}$.
This process continues until the trajectory is completed and the continuum robot has successfully steered into the target vessel (Fig.~\ref{fig_control_schematic}).
During tracking at index $k$, the feedforward control uses the precomputed input from the plan: 
$\mathbf{u}_{\,t}^{\text{FF}}
= 
\mathbf{u}_k$.

\subsection{Feedback Control} \label{sec_feedback_control}

\subsubsection*{Contact-Aware Robot Jacobian} 

Incremental changes in joint configuration $\Delta \mathbf{q}$ relate approximately to changes in tip state $\Delta \mathbf{p}$ through a first-order linear relationship:
\begin{equation}
    \Delta \mathbf{p} \approx \mathbf{J} \, \Delta \mathbf{q}.
    \label{eq_derivlinear_relation}
\end{equation}
After computing the contact-aware motion plan, while still offline, we iterate through each index $k$ and compute the contact-aware robot Jacobian $\mathbf{J}^{\Delta}$ by perturbing each joint and measuring the resulting change in tip state:
\begin{equation}
    \mathbf{J}^{\Delta} \;=\;
    \begin{bmatrix}
        \frac{\Delta x}{\Delta q_{\mathrm{a}}} & \frac{\Delta x}{\Delta q_{\mathrm{p}}} & \frac{\Delta x}{\Delta q_{\mathrm{d}}} \\
        \frac{\Delta z}{\Delta q_{\mathrm{a}}} & \frac{\Delta z}{\Delta q_{\mathrm{p}}} & \frac{\Delta z}{\Delta q_{\mathrm{d}}} \\
        \frac{\Delta \phi}{\Delta q_{\mathrm{a}}} & \frac{\Delta \phi}{\Delta q_{\mathrm{p}}} & \frac{\Delta \phi}{\Delta q_{\mathrm{d}}} \\ 
    \end{bmatrix} .
    \label{eq_perturbed_Jac}
\end{equation}
The resulting sequence of contact-aware Jacobians $\mathbf{J}^{\ \ \Delta}_{0:{\tau-1}}$
provides the joint-to-task-space mapping for all plan indices during feedback control.
The Jacobian and its condition number vary significantly with robot pose and contact location. 
Although numerically approximate, this Jacobian provides sufficient accuracy for closed-loop trajectory tracking

\subsubsection{PD Control}
We use the damped pseudo-inverse of the contact-aware robot Jacobian in a proportional-derivative (PD) feedback controller:
\begin{equation}
    \mathbf{u}_{\,t}^{\text{FB}} =
    (\mathbf{J}^{\Delta}_{\,k})^{\dagger} \,
    \left( 
    \mathbf{K}_p \mathbf{e}_{k,t} + 
    \mathbf{K}_d \dot{\mathbf{e}}_{k,t}
    \right)
    \label{eq_FB_control}
\end{equation}
where $\mathbf{K}_p = \text{diag}(K_p^x, K_p^z, K_p^\phi)$ weights positional (mm) and orientation (rad) errors in $\mathbf{e}_{k,t}$
and 
$\mathbf{K}_d = \text{diag}(K_d^x, K_d^z, K_d^\phi)$ scales the error derivative $\dot{\mathbf{e}}_{k,t}$.

\section{Continuum Robot Hardware} \label{sec_robot_design}

\begin{figure}[t!]
\centerline{\includegraphics[width=1.0\linewidth]{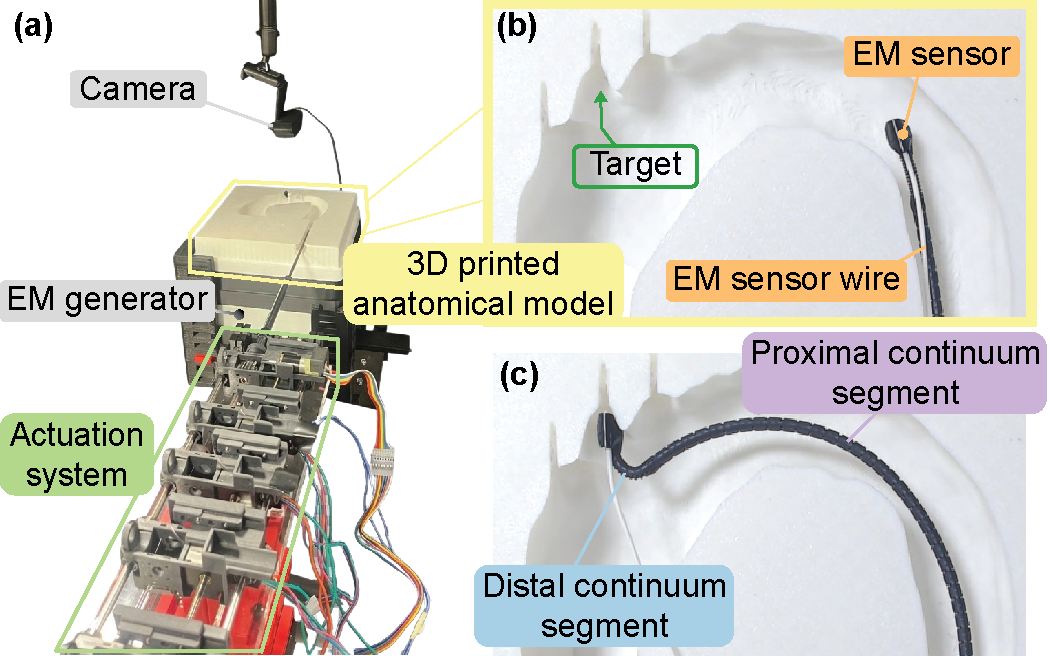}}
\vspace{-8pt}
\caption{Continuum robot hardware used to evaluate the contact-aware planning and control framework. 
(a) The actuation system (green) contains five stepper motors: four for antagonistic tendon actuation and one that drives insertion.
A camera positioned above the workspace emulates planar fluoroscopy commonly used during neurointerventional procedures. 
(b) A 3D-printed anatomical model generated from a patient CT scan sits above the electromagnetic (EM) generator used together with the EM tip sensor to track tip pose $\mathbf{p}_{t}^{\ast}$.
(c) The notched continuum robot consists of proximal and distal bending segments (parameters are provided in Table~\ref{table_notch_params}), enabling navigation around the aortic arch toward the target vessel. 
}
\label{fig_hardware_setup}  
\end{figure}

We designed and fabricated an antagonistically driven notched continuum robot and actuation system to evaluate the proposed planning and control framework (Fig.~\ref{fig_hardware_setup}).
Table~\ref{table_notch_params} summarizes the geometric parameters of the notched continuum segments that enable navigation across type~I, II, and III aortic arches while reducing joint coupling and minimizing the actuation forces required for the distal continuum segment~\cite{mangan2025serially}.
We 3D printed the continuum robot using a Form~4 printer (Tough 1500 Resin V2, Formlabs Inc.).
The design integrates four tendon routing channels for antagonistic actuation with 0.28 mm diameter nitinol tendons (NW-011-72-05, Small Parts) adhered with UV cure adhesive (AA3936, Loctite).
The tendons are routed back through a rigid multi lumen shaft to the actuation system.

The actuation system (green in Fig.~\ref{fig_hardware_setup}) drives five degrees of freedom. 
Each tendon connects to an individual stepper motor (Nanotec LA281S10-A-THCA) and all are mounted to a stationary lead screw (Nanotec SCREW-AAA-THCA-1000) to provide precise insertion increments.
An additional stepper motor and lead screw assembly provide system insertion.

\section{Experiments and Results}
\begin{figure}[t!]
\centerline{\includegraphics[width=1.0\linewidth]{Figures/fig_exp_sim-007.png}}
\vspace{-8pt}
\caption{Independent of controller choice, a reference trajectory must respect the robot kinematics and contact constraints. 
As an example, a model-free~\cite{yip2014model} controller (a) failed when tracking a centerline trajectory with kinematically infeasible pose targets, but (b) succeeded when the same controller tracked the trajectory from the proposed contact-aware planner.}
\label{fig_exp_Kin_Feas}  
\end{figure}
\begin{figure*}[t]
\centerline{\includegraphics[width=1.0\linewidth]{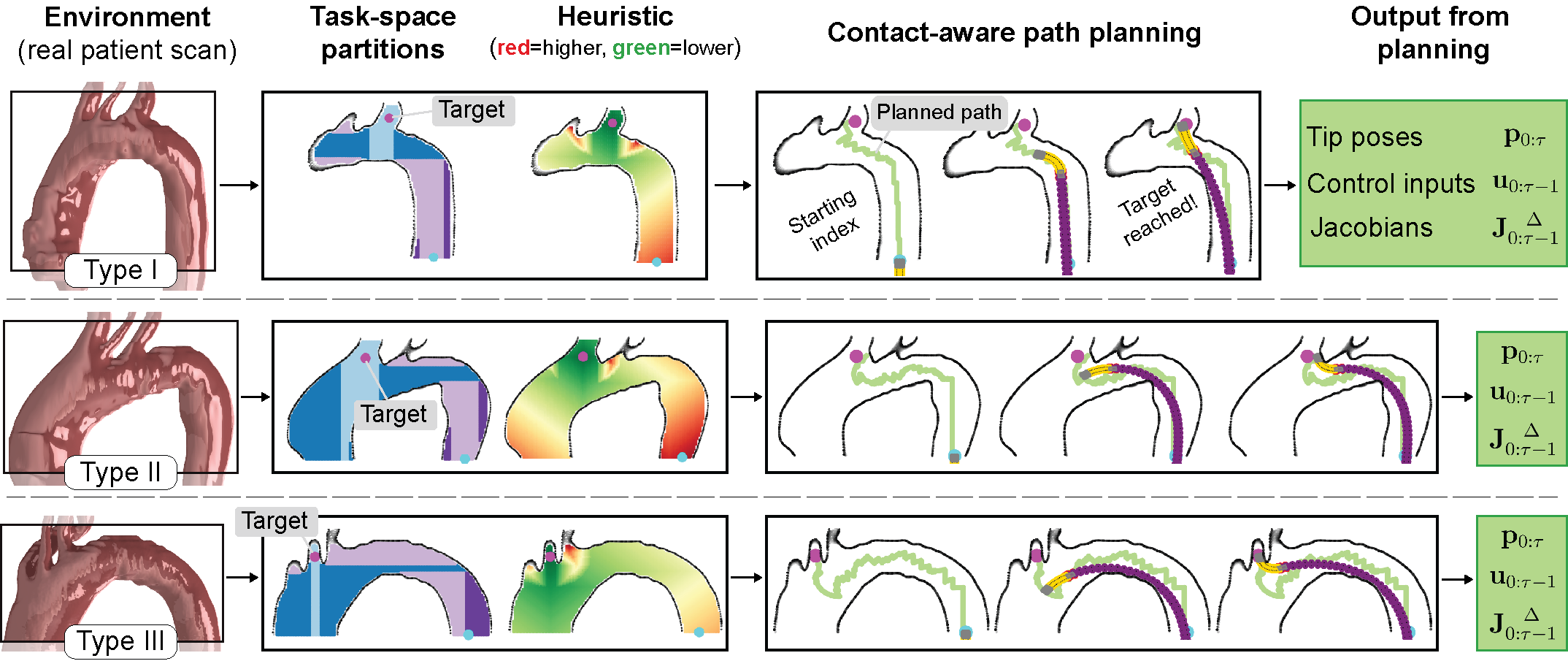}}
\vspace{-8pt}
\caption{Planning across three anatomical environments. The algorithm begins with dividing a (real) patient scan into task-space partitions and creating a heuristic (Section~\ref{sec_heuristic}). 
Then the search-based planner minimizes end of continuum segment (ECS) contact with the environment.
The planner outputs tip poses $\mathbf{p}_{0:\tau}$, control inputs $\mathbf{u}_{0:\tau-1}$, and contact-aware Jacobians $\mathbf{J}^{\ \Delta}_{0:\tau-1}$ which are used when navigating in hardware.
The proposed contact-aware planning method is robust to multiple aortic arch environments.}
\label{fig_exp_PlanningAll}  
\end{figure*}

In this section, we describe the evaluation of the proposed contact-aware planning and control framework.
We first detail the experimental setup, including the anatomical environments created from patient scans.
We then demonstrated the need for contact-aware planning over a naive centerline approach and showed the robustness of our method across various environments.
Hardware experiments validated the approach through successful navigation across three anatomical environments (Supplementary Video S1).
Lastly, we conducted ablation studies to assess the importance of (1) allowing different types of contact and (2) using contact-aware Jacobians with closed-loop control.

\subsection{Experiment Setup, Environments, and Metrics} \label{sec_exp_setup}
We evaluated the proposed contact-aware planning and control framework in anatomical environments reconstructed from patient scans.
Guided by morphology characterization~\cite{inancc2017effect} and in consultation with clinical collaborators, we selected three representative aortic arch types (type I, type II, and type III) from a segmented dataset~\cite{radl2022avt}. 
The arch types spanned the range of procedure difficulty, from more straightforward (type I) to more challenging curvatures (type III).
We restricted our study to arches that only required planar steering.
We exported each model from 3D Slicer~\cite{kikinis20133d} as an STL file and used the meshes to (1) construct the environment point clouds $\Omega$ for contact-aware planning and (2) fabricate the 3-D printed test environments (Bambu Lab X1C).
For hardware experiments, we left the superior portion of each model open to allow for visualization similar to fluoroscopy (Fig.~\ref{fig_hardware_setup}).
The imaging was not used in control. 
We tracked robot tip pose ($\mathbf{p}_{t}^{\ast}$) with an electromagnetic (EM) sensor (Aurora 6DOF FlexTube Type 2) and field generator (Northern Digital Inc.) for closed-loop control. 

The proposed framework was evaluated using three primary metrics:
(1) the success rate in hardware trials (percentage of trials that reached the target pose without tip contact) 
(2) the completion time, 
(3) the distance ($\overline{e}_{\mathrm{avg}}$) between the planned path and measured tip position.
Although $\mathbf{p}_{k}, \mathbf{p}_{t}^{\ast} \in \mathbb{R}^{3}$ included planar position and orientation, we evaluated tracking error using position only.
At time $t$ we computed the distance between the measured tip pose $\mathbf{p}_{t}^{*}$ and planned trajectory $\mathbf{p}_{0:\tau}$ as
\begin{equation} \label{eq_measure_dist_fromPlannedPath}
    e_t =
    \min_{\substack{\mathbf{p}_{k} \in \mathbf{p}_{0:\tau}}} 
    \dist{\mathbf{p}_{t}^{\ast}}{\mathbf{p}_{k}} \\.
\end{equation}
We collected a sequence $e_{0:T}$ over each trial, where $T$ denoted the time at which the tip reached the goal set $\mathbf{p}_{t}^{\ast} \in \mathcal{G}$.
The mean tracking error for a trial was: 
\begin{equation} \label{eq_measure_dist_fromPlannedPath_mean}
    \overline{e} = \frac{1}{T+1} \sum_{t=0}^{T} e_t ,
\end{equation}
and we reported the average across three trials as $\overline{e}_{\mathrm{avg}}$.
\subsection{Planning Performance}
\subsubsection*{Importance of Contact-Aware Planning} \label{sec_exp_kin_infeasible}

We first demonstrated the necessity of contact-aware planning by comparing it against a naive trajectory that neglects the kinematics of the robot.
A model-free controller~\cite{yip2014model} was implemented with two different target trajectories: (a) naive centerline which ignores the robot kinematics and contact constraints and (b) the trajectory generated by the proposed contact-aware planner.
Even with ideal sensing and perfect Jacobian estimation in simulation, the naive centerline trajectory and model-free control failed in certain environments (Fig.~\ref{fig_exp_Kin_Feas}).
With the naive centerline trajectory, the robot initially tracked well but then reached a kinematically infeasible pose target, causing the robot to stop moving and fail (Fig.~\ref{fig_exp_Kin_Feas}(a)).
In contrast, when the target tip trajectory was generated by the proposed contact-aware planner, all poses were feasible, and the robot successfully reached the goal using same controller (Fig.~\ref{fig_exp_Kin_Feas}(b)).
\subsubsection*{Planning in Anatomical Environments} \label{sec_exp_planning}

\begin{table} [t] 
\centering
\vspace{-8pt}
\caption{Contact-Aware Planning Results} \label{table_planning_nominal_results}
\begin{center}
\begin{tabular}{|c|c|c|c|}
\hline
\textbf{Environment} & 
\textbf{\makecell{Modification*}} & 
\textbf{\makecell{Expands}} & 
\textbf{\makecell{Time (s)}} \\ 
\hline
Type I & \cellcolor[HTML]{a6cee3}None & 26,916 & 1531 \\
\hline
Type II & \cellcolor[HTML]{a6cee3}None & 7,337 & 298 \\
\hline
Type III & \cellcolor[HTML]{a6cee3}None & 42,774 &  838 \\
\hline
Type III & \makecell{Without partition-dep. \\ action space} & 45,965 & 1096 \\
\hline
Type III & \makecell{Removed body \\ contact penalty$^\ominus$} & 1,373 & 50 \\
\hline
Type II & \makecell{Removed ECS and \\ body contact penalty$^\otimes$} & 2,166 & 119 \\
\hline
\end{tabular}
\end{center}
\raggedright
\par Modification* indicates deviations from the proposed method described in Section~\ref{sec_contact_planner}.
\par $^\ominus$ resulted in more body contact in hardware experiment
\par $^\otimes$ failed in hardware experiment
\end{table}

We evaluated the robustness and performance of the contact-aware motion planner on the three anatomical environments.
The search-based planner successfully converged in all three arch types (Fig.~\ref{fig_exp_PlanningAll}). 
All algorithms were implemented in Python and executed on a 64-bit Ubuntu Linux workstation with an Intel Core~i7-8700 CPU @3.20~GHz (6 cores, 12 threads), 64~GB RAM. 

The planner converged in under 30 minutes for all three anatomical environments (Table~\ref{table_planning_nominal_results}). 
The type III aortic arch plan took longer to converge than the type II plan (838~s vs. 298~s respectively), consistent with the increased path difficulty observed in clinical practice.
The type I arch required the longest convergence time (1531~s), primarily due to the longer proximal continuum segment and ECS contact penalty in Eq.~\eqref{eq_contact_cost}.
Avoiding ECS contact during actuation of the proximal segment is more difficult because the same continuum robot (notch parameters in Table~\ref{table_notch_params}) and planning parameters (e.g., $\mathbf{u}_{\mathrm{step}}$) were used across all arch types and was not optimized for the type I geometry. 
We note that path smoothness depends on the control input magnitude $\mathbf{u}_{\mathrm{step}}$ (Section~\ref{sec_dynamic_action_space}) used during planning: smaller increments produced smoother trajectories at the expense of convergence time.
For all plans we used $\mathbf{u}_{\mathrm{step}} = [1.0, 0.1, 0.1]^{\top}$ mm as this was a realistic action increment for our actuation system to command.

To quantify the computational cost of the proposed contact penalties, we performed an ablation study. 
Removing the body contact penalty in a type III arch reduced convergence time by more than $15 \times$; however, this speedup primarily resulted from a $31 \times$ reduction in node expansions, rather than from eliminating the contact cost computation itself.
Similarly, removing the ECS and body contact penalties while planning a type II arch resulted in a $2.5 \times$ reduction in convergence time while evaluating about $3 \times$ less nodes.
Section~\ref{sec_exp_integrated_sys} demonstrated the practical benefits of including these contact costs in the integrated system experiments.
Lastly, we found that using the partition-dependent action space over a static action space decreased planning time by $23\%$ (Table~\ref{table_planning_nominal_results}).

\begin{figure}[t]
\centerline{\includegraphics[width=1.0\linewidth]{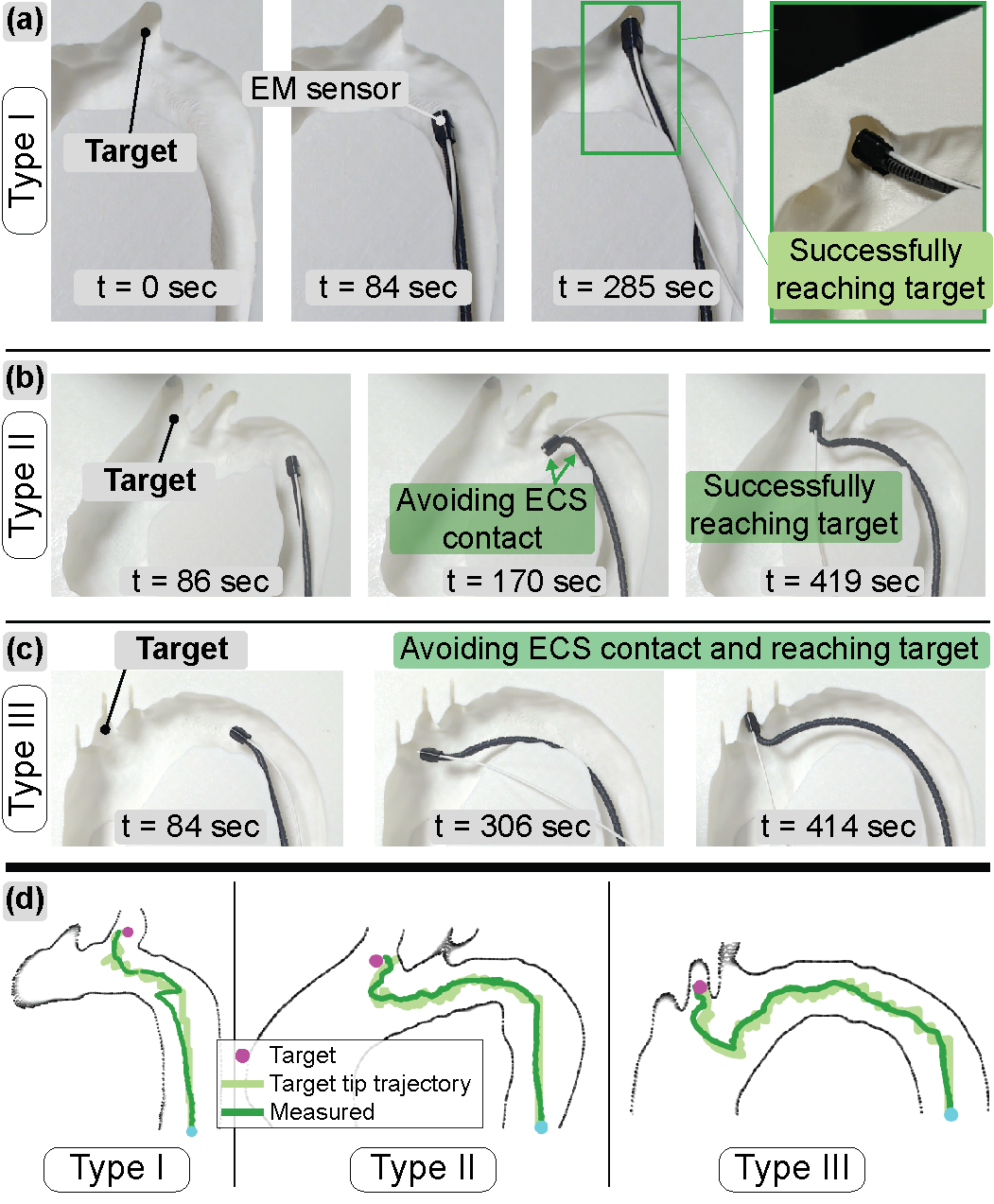}}
\vspace{-2pt}
\caption{
Hardware experiments evaluated the proposed contact-aware planning and control framework in three anatomical environments: 
(a) type I, (b) type II, and (c) type III aortic arches.
The continuum robot avoided contact at the end-of-continuum-segment (ECS) and reached the target location, through which a guidewire or soft growing robot can be deployed. 
(d) Measured tip position (dark green) and target trajectory (light green) from contact-aware planning for one trial from each arch type (I, II, and III)
We conducted three trials in each environment (nine total trials), and the robot successfully reached the target vessel in all trials.
}
\label{fig_exp_hardware_all_exampletrial}  
\end{figure}

\subsection{Hardware Experiments: Integrated Planning and Control Pipeline} \label{sec_exp_integrated_sys}
The robot successfully reached the target tip pose in all trials while avoiding distal tip contact for all environments in hardware experiments (Fig.~\ref{fig_exp_hardware_all_exampletrial}).
We repeated each experiment three times per environment (nine total trials). 
Figure~\ref{fig_exp_hardware_all_exampletrial}(d) illustrates the position of both the target trajectory $\mathbf{p}_{0:\tau}$ (light green) and measured tip pose $\mathbf{p}_{0:T}^{\ast}$ (dark green) of a representative trial for each arch type.

As described in Section~\ref{sec_exp_setup}, the main metrics assessed for each arch type were the average tracking error and completion time. As a rough order-of-magnitude comparison, we note that clinical stroke intervention studies often cite a 60 minute window as an important benchmark associated with improved patient outcomes~\cite{spiotta2014golden}.
For the type I arch, the average tracking error across three trials was $\overline{e}_{\mathrm{avg}} = 1.9 \pm 0.5$~mm.
The total time required to go from scan to successful navigation was 1786~s (29.8 minutes) -- 1531~s (Table~\ref{table_planning_nominal_results}) for planning and $255 \pm 55$~s for navigation. 
For the type II arch, the average tracking error was $\overline{e}_{\mathrm{avg}} = 1.2 \pm 0.1$~mm, with an average completion time of 731~s (12.2~min) -- 298~s for planning and $433 \pm 19$~s for navigation.
The type III arch, traditionally the hardest for neurointerventionalists to navigate~\cite{knox2020impact}, had an average tracking error of $\overline{e}_{\mathrm{avg}} = 1.7 \pm 0.2$~mm. The total time was 1267~s (21.1~min) -- 839~s for planning and $428 \pm 14$~s for navigation.
Across all experiments, the robot tracked the planned path closely and avoided unintended distal tip contact throughout execution.

\subsection{Ablation Studies}
We isolated key components of the proposed planning and control framework to highlight their importance.
Specifically, we evaluated the effects of (1) avoiding ECS contact during planning, (2) penalizing body contact, and (3) incorporating closed-loop feedback during execution.
\begin{figure}[t]
\centerline{\includegraphics[width=1.0\linewidth]{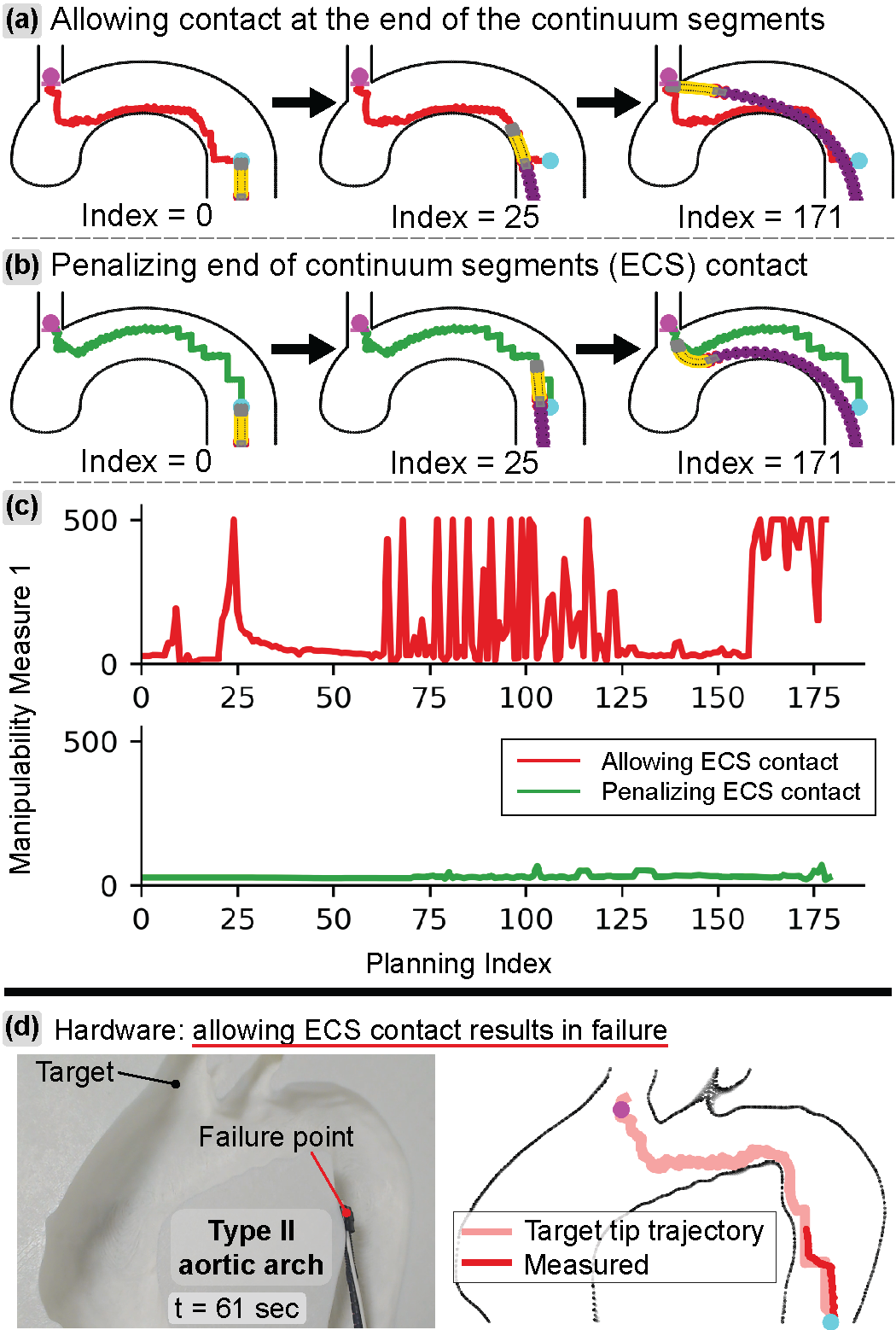}}
\vspace{-8pt}
\caption{Comparison of contact-aware plans: 
(a) allowing contact at the end of the continuum segments (ECS), shown in red, and 
(b) penalizing ECS contact, shown in green. 
(c) Manipulability measure~1 ($\mu_1$)~\cite{lynch2017modern} along each trajectory. 
Lower $\mu_1$ values indicated a well-conditioned Jacobian, while larger $\mu_1$ indicated a loss of rank. 
(d) Following a contact-aware plan (light red) which allowed ECS contact resulted in failure when testing on hardware in the type II aortic arch. 
}
\label{fig_exp_ECScontact}  
\end{figure}
We first quantified how avoiding ECS contact in a simulation environment affected controllability and then validated the effect on hardware.
After generating plans with and without ECS contact penalty, we estimated manipulability~\cite{lynch2017modern} at each configuration in the planned trajectory as:
\begin{equation}
    \mu_1(J) = \sqrt{\frac{\lambda_{max}(\mathbf{J}\mathbf{J}^\top)} {\lambda_{min}(\mathbf{J}\mathbf{J}^\top)}}
    \label{eq_manipulability_measure1}
\end{equation}
where $\lambda_{max}$ and $\lambda_{min}$ are the largest and smallest eigenvalues of matrix $\mathbf{J}\mathbf{J}^\top$.
A low value of $\mu_1$ indicates a full-rank, well-conditioned Jacobian, while $\mu_1$ approaching infinity corresponds to a singularity.
We computed the Jacobian numerically by perturbing $\mathbf{q}$ at each step and tracking the resulting tip displacement similar to Eq.~\eqref{eq_perturbed_Jac}.
For visualization we clipped $\mu_1$ at 500.

The plan that did not penalize ECS contact exhibited substantially higher $\mu_1$ values throughout the trajectory, indicating less favorable manipulability (Fig.~\ref{fig_exp_ECScontact}(a-c)).
We then evaluated the performance in hardware on the type II aortic arch.
The plan without ECS contact penalty (light red in Fig.~\ref{fig_exp_ECScontact}(d)) produced tip contact and caused the robot to become stuck while trying to command into the vessel wall.
In hardware experiments, if the target trajectory was near an obstacle and the obstacle was misaligned during registration, it was impossible to reach the desired tip pose.
In contrast, penalizing ECS contact during planning preserved controllability and enabled successful navigation in all hardware trials (Fig.~\ref{fig_exp_hardware_all_exampletrial}).

\begin{figure}[t]
\centerline{\includegraphics[width=1.0\linewidth]{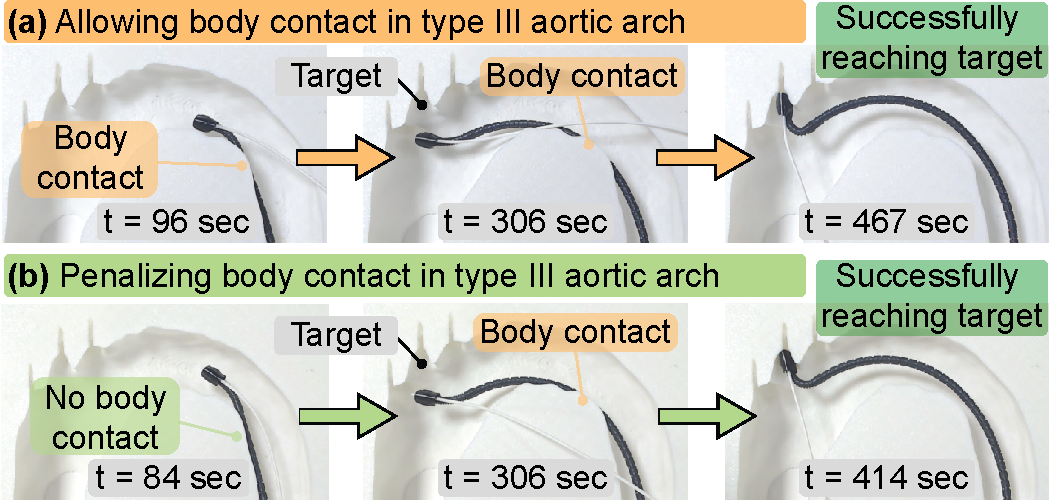}}
\vspace{-2pt}
\caption{
Successfully navigating type III aortic arch with plans that (a) allowed body contact and (b) penalized body contact.
The proposed planner reduced body contact during execution (Table~\ref{table_bodycontact_hardware_results}).
}
\label{fig_exp_hardware_BODYcontact}  
\end{figure}
\begin{table}
\vspace{-8pt}
\caption{Hardware Results: Planning with Body Contact Cost}
\label{table_bodycontact_hardware_results}
  \centering
  \begin{center}
  \begin{tabular}{ccc}
    \toprule
           Method & 
           \makecell{Body Contact \\ Time} &
           \makecell{Completion \\ Time} \\
           \midrule
           \multicolumn{3}{l}{\underline{\textit{With Body Contact Penalty (proposed method)}}} \\
           Trial 1 & 343 s  & 430 s  \\
           Trial 2 & 308 s  & 414 s \\
           Trial 3 & 333 s  & 441 s  \\
           \textbf{Average} & \textbf{328 s} & \textbf{428 s}\\
           \addlinespace[2pt]
           \multicolumn{3}{l}{\underline{\textit{No Body Contact Penalty}}} \\
           Trial 1 & 371 s & 493 s \\
           Trial 2 & 366 s & 467 s \\
           Trial 3 & 310 s & 405 s \\
           \textbf{Average} & \textbf{349 s} & \textbf{455 s} \\
  \end{tabular}
  \end{center}
  \vspace{2mm}
  \raggedright
\end{table}
We next characterized the effect of the body contact penalty by running the planner without the body contact cost (while retaining the ECS penalty).
We performed three trials and visually measured the time in contact with the environment (Fig.~\ref{fig_exp_hardware_BODYcontact}).
Planning with the body contact penalty reduced average body contact time by $21$~s in hardware and reduced average completion time by $27$~s (Table~\ref{table_bodycontact_hardware_results}).
In hardware, the environment is not perfectly aligned with the theoretical configuration assumed during planning, causing the estimated contact-aware Jacobians $\mathbf{J}^{\ \Delta}_{0:\tau-1}$ to more closely approximate the true robot Jacobian when body contact is absent.
We therefore hypothesize that reducing body contact improved agreement between the planned and actual interaction conditions, which likely contributed to the reduced average completion time (Table~\ref{table_bodycontact_hardware_results}).
When body contact is present, discrepancies between theoretical and actual environment geometry amplify Jacobian error, degrading control accuracy and increasing completion time.

\begin{figure}[t]
\centerline{\includegraphics[width=1.0\linewidth]{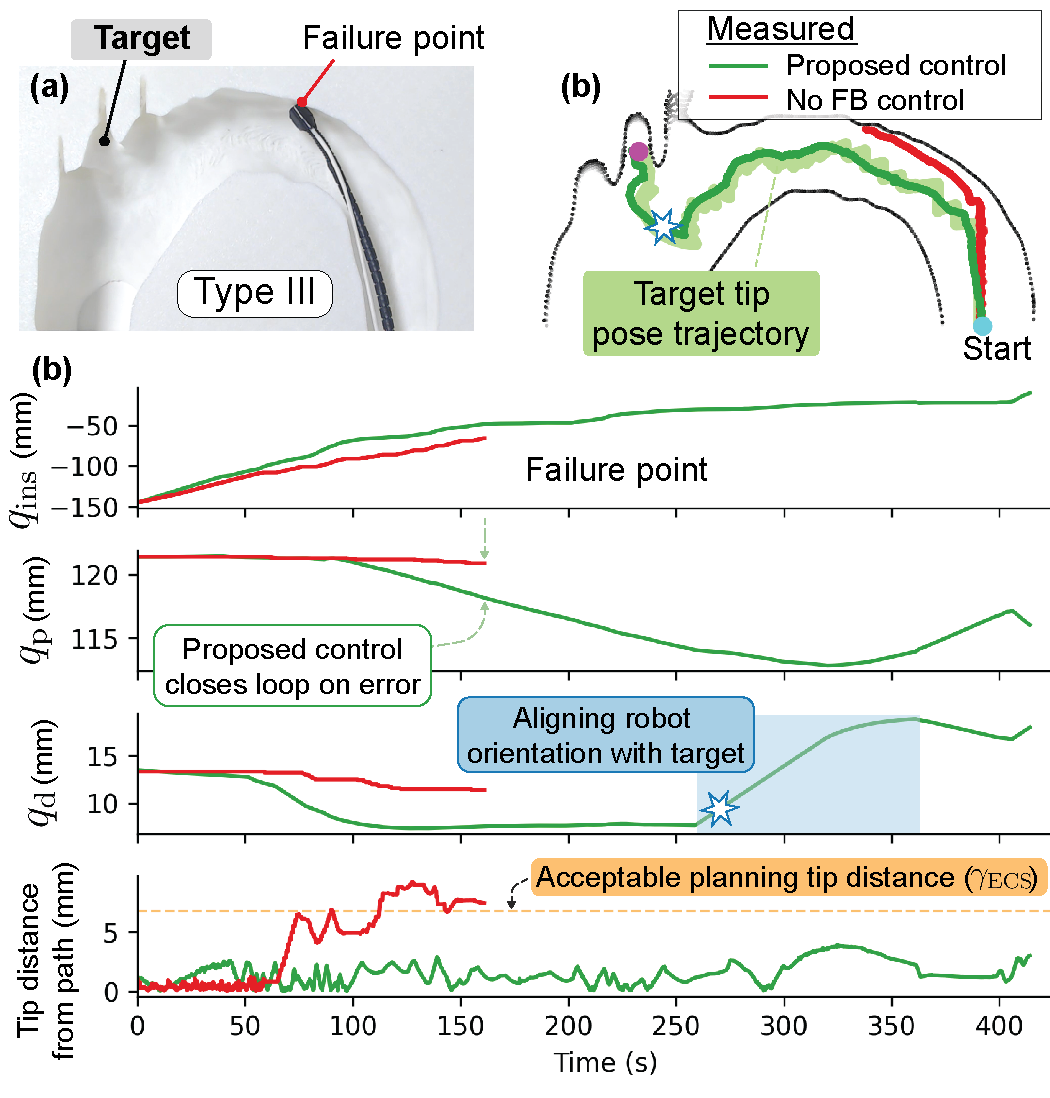}}
\vspace{-8pt}
\caption{Closed-loop feedback was required for successful execution. (a) Open-loop (red) resulted in tip contact and large error from the planned trajectory while closed-loop (green) maintained tip position near the planned trajectory and successfully navigated to the target. 
(b) The difference between the open-loop and closed loop $\mathbf{q}$ values showed how critical closed-loop control was.
The tip distance from the path was above $\gamma_{\text{ECS}}$ for the open-loop trial and remained under $\gamma_{\text{ECS}}$ for the closed loop trial.
}
\label{fig_exp_FB}  
\end{figure}

Finally, we evaluated the importance of closed-loop feedback in type III trials.
Open-loop execution of $\mathbf{q}_{0:\tau}$ from the planner resulted in tip contact with the environment and failing to reach the target (Fig.~\ref{fig_exp_FB}(a-b)).
In contrast, closed-loop control updated the joint values $\mathbf{q}$ at each time step to track the desired tip pose and compensate for model inaccuracy, actuation error, and environment misalignment.
This enabled the tip distance to the path to remain below $\gamma_{\text{ECS}}$ (Fig.~\ref{fig_exp_FB}(c)).

\section{Conclusion}
Autonomous navigation of continuum robots in confined anatomical environments is an important yet challenging problem, as unavoidable contact with surrounding structures can degrade manipulability, induce singularities, and introduce safety risks.
This work presents a unified planning and control framework that enables trajectories that discourage hazardous contact while allowing benign contact when necessary.
The planner produces kinematically feasible trajectories that discourage unfavorable contact and provides contact-aware Jacobians that enable trajectory tracking with minimal sensing during hardware execution.
Experiments with a multi-segment continuum robot in anatomical models (from patient scans) demonstrated reliable navigation of type I, type II, and type III aortic arches with consistent target reaching, low tracking error, and avoidance of distal tip contact.
Ablation studies showed that avoiding contact at the end of the continuum segments improved manipulability and prevented hardware failures.
Future work will concentrate on extending the current 2D formulation to 3D, further reducing computation time for planning, and improving execution time in hardware.
Future work can also include the integration of state estimation techniques to infer robot-environment interaction forces and to inform both the planner and controller, further reducing the risk of tissue injury in clinical settings.

\appendices

\section{Contact-Aware Motion Model Derivatives} \label{sec_derivatives}

This appendix provides the analytical derivative formulations for the contact-aware motion model optimization in ~\eqref{eq_motionmodel_opt}. The optimization operates in a high-dimensional space with $m_p + m_d = 46$ curvature variables, making analytical derivatives essential for computational efficiency.

\subsection{Analytical Derivative Approach}

The key insight enabling efficient analytical derivatives is the forward propagation of gradients through the kinematic chain using the chain rule. For a transformation chain $T_{\text{final}} = T_1 \cdot T_2 \cdot \ldots \cdot T_n$, the derivative with respect to parameter $\kappa_j$ affecting transformation $T_k$ is:
\begin{equation} \label{eq_chain_rule_transform}
\frac{\partial T_{\text{final}}}{\partial \kappa_j} = T_1 \cdot \ldots \cdot T_{k-1} \cdot \frac{\partial T_k}{\partial \kappa_j} \cdot T_{k+1} \cdot \ldots \cdot T_n
\end{equation}
This allows us to compute all gradients in a single forward pass, avoiding the computational overhead of finite difference approximations.

\subsection{Objective Gradient}

The objective function minimizes the sum of squared curvatures:
\begin{equation} \label{eq_obj_gradient}
\frac{\partial}{\partial \boldsymbol{\kappa}} \sum_{i=1}^{m_p+m_d} \kappa_i^2 = 2 \boldsymbol{\kappa}
\end{equation}
where $\boldsymbol{\kappa} \in \mathbb{R}^{m_p+m_d}$ is the curvature vector.

\subsection{Constraint Jacobians}

The optimization has three constraints (two equality constraints for tendon lengths and one inequality constraint for environment clearance), resulting in a constraint Jacobian $\mathbf{J}_{\text{C}} \in \mathbb{R}^{3 \times (m_p + m_d)}$.

\subsubsection{Transformation Matrix Derivatives}

The foundation for all constraint gradients is computing how robot points change with curvature. Each continuum segment uses a transformation matrix from Eq.~\eqref{eq_constcurv_Trans}. For a transformation $T_{j_{pc}}^{j+1_p}(\kappa_j)$ representing the $j$-th segment, the derivative with respect to its curvature $\kappa_j$ is:

\begin{align} \label{eq_constcurv_Tderiv}
&\frac{\partial T_{j_{pc}}^{j+1_p}(\kappa_j)}{\partial \kappa_j} = \\ 
&\begin{bmatrix}
   -l_j \sin(\kappa_j l_j) & 0 & l_j \cos(\kappa_j l_j) & 
   \frac{l_j \sin(\kappa_j l_j)}{\kappa_j} - \frac{1 - \cos(\kappa_j l_j)}{\kappa_j^2} \\
   0 & 0 & 0 & 0 \\
   -l_j \cos(\kappa_j l_j) & 0 & -l_j \sin(\kappa_j l_j) & \frac{l_j \cos(\kappa_j l_j)}{\kappa_j} - \frac{\sin(\kappa_j l_j)}{\kappa_j^2} \\
   0 & 0 & 0 & 0 \\
\end{bmatrix}
\end{align}
where $l_j$ is the arc length of segment $j$ (either $l_i^p$ for proximal or $l_i^d$ for distal segments).

\subsubsection{Point Position Derivatives via Chain Rule}

For a point $\mathbf{p}_{s}^{i_p} \in \mathbb{R}^3$ at frame $\{i_p\}$ expressed in space frame $\{s\}$, the transformation chain is:
\begin{equation} \label{eq_transform_chain}
T_s^{i_p}(\boldsymbol{\kappa}) = T_s^{0_{pc}} \cdot T_{0_{pc}}^{1_p}(\kappa_0) \cdot T_{1_{p}}^{1_{pc}} \cdot T_{1_{pc}}^{2_p}(\kappa_1) \cdot \ldots \cdot T_{i-1_{pc}}^{i_p}(\kappa_{i-1})
\end{equation}

The derivative with respect to curvature $\kappa_j$ follows the chain rule:
\begin{align} \label{eq_partial_T_Kj}
&\frac{\partial T_s^{i_p}(\boldsymbol{\kappa})}{\partial \kappa_j} = \\
&\begin{cases}
\mathbf{0} & \text{if } j \geq i \\
T_s^{0_{pc}} \cdot \ldots \cdot \frac{\partial T_{j_{pc}}^{j+1_p}(\kappa_j)}{\partial \kappa_j} \cdot T_{j+1_p}^{j+1_{pc}} \cdot \ldots \cdot T_{i-1_{pc}}^{i_p}(\kappa_{i-1}) & \text{if } j < i
\end{cases}
\end{align}

The position of point $\mathbf{p}_{s}^{i_p}$ is extracted from the translation component of $T_s^{i_p}$:
\begin{equation}
\mathbf{p}_{s}^{i_p} = \left[ T_s^{i_p} \right]_{1:3, 4}
\end{equation}
where $[\cdot]_{1:3, 4}$ extracts the first three elements of the fourth column (translation vector).

Therefore, the derivative of the point position is:
\begin{equation} \label{eq_point_derivative}
\frac{\partial \mathbf{p}_{s}^{i_p}}{\partial \kappa_j} = \left[ \frac{\partial T_s^{i_p}(\boldsymbol{\kappa})}{\partial \kappa_j} \right]_{1:3, 4}
\end{equation}

For points offset from the centerline (e.g., left/right cable attachment points), additional transformations $T_L$ and $T_R$ are applied:
\begin{equation}
\frac{\partial (T_s^{i_p} \cdot T_L \cdot [0, 0, 0, 1]^T)}{\partial \kappa_j} = \frac{\partial T_s^{i_p}}{\partial \kappa_j} \cdot T_L \cdot [0, 0, 0, 1]^T
\end{equation}

\subsubsection{Tendon Length Constraint Jacobian}

The tendon length constraints enforce consistency between actuated lengths $q_{p,t}$, $q_{d,t}$ and calculated lengths. The calculated proximal tendon length is:
\begin{equation}
l_{\text{ten,calc}}^p = c_p(m_p-1) + \sum_{i=0}^{m_p-2} \| \mathbf{p}_{s}^{i+1_{pc}} - \mathbf{p}_{s}^{i+1_p} \|
\end{equation}
where $c_p$ is the uncut height (constant), and the sum represents cable lengths between consecutive segments along the left side of the tube.

The constraint is $g_p(\boldsymbol{\kappa}) = q_{p,t} - l_{\text{ten,calc}}^p = 0$. Its gradient is:
\begin{equation} \label{eq_tendon_length_grad}
\frac{\partial g_p}{\partial \kappa_j} = -\frac{\partial l_{\text{ten,calc}}^p}{\partial \kappa_j} = -\sum_{i=0}^{m_p-2} \frac{\partial \| \mathbf{p}_{s}^{i+1_{pc}} - \mathbf{p}_{s}^{i+1_p} \|}{\partial \kappa_j}
\end{equation}

For each segment, let $\mathbf{v}_i = \mathbf{p}_{s}^{i+1_{pc}} - \mathbf{p}_{s}^{i+1_p}$ be the vector between points. The derivative of its norm is:
\begin{equation}
\frac{\partial \| \mathbf{v}_i \|}{\partial \kappa_j} = \frac{\mathbf{v}_i^T}{\| \mathbf{v}_i \|} \cdot \frac{\partial \mathbf{v}_i}{\partial \kappa_j} = \hat{\mathbf{v}}_i^T \left( \frac{\partial \mathbf{p}_{s}^{i+1_{pc}}}{\partial \kappa_j} - \frac{\partial \mathbf{p}_{s}^{i+1_p}}{\partial \kappa_j} \right)
\end{equation}
where $\hat{\mathbf{v}}_i = \mathbf{v}_i / \| \mathbf{v}_i \|$ is the normalized direction vector.

Combining all segments:
\begin{equation} \label{eq_tendon_grad_final}
\frac{\partial g_p}{\partial \kappa_j} = -\sum_{i=0}^{m_p-2} \hat{\mathbf{v}}_i^T \left( \frac{\partial \mathbf{p}_{s}^{i+1_{pc}}}{\partial \kappa_j} - \frac{\partial \mathbf{p}_{s}^{i+1_p}}{\partial \kappa_j} \right)
\end{equation}

The distal tendon length constraint gradient follows the same form with $m_d$ segments and $c_d$.

\subsubsection{Environment Constraint Jacobian}

The environment constraint ensures all robot points maintain minimum distance $d_{\text{thr}}$ from anatomy. For each point $\mathbf{p}_k \in \mathcal{R}(q_{\mathrm{a},k},\boldsymbol{\kappa})$, we compute the distance to the nearest anatomy point:
\begin{equation}
d_k(\boldsymbol{\kappa}) = \min_{\mathbf{a} \in \Omega} \| \mathbf{p}_k(\boldsymbol{\kappa}) - \mathbf{a} \|
\end{equation}

The constraint is $g_{\text{env},k}(\boldsymbol{\kappa}) = d_k(\boldsymbol{\kappa}) - d_{\text{thr}} \geq 0$ for all points $k$.

Let $\mathbf{a}_k^*$ be the nearest anatomy point to $\mathbf{p}_k$, and $\mathbf{d}_k = \mathbf{p}_k - \mathbf{a}_k^*$ be the vector from anatomy to robot point. The distance derivative is:
\begin{equation}
\frac{\partial d_k}{\partial \kappa_j} = \frac{\partial \| \mathbf{d}_k \|}{\partial \kappa_j} = \hat{\mathbf{d}}_k^T \frac{\partial \mathbf{p}_k}{\partial \kappa_j}
\end{equation}
where $\hat{\mathbf{d}}_k = \mathbf{d}_k / \| \mathbf{d}_k \|$ is the normalized direction vector.

Therefore, the constraint gradient is:
\begin{equation} \label{eq_env_grad}
\frac{\partial g_{\text{env},k}}{\partial \kappa_j} = \hat{\mathbf{d}}_k^T \frac{\partial \mathbf{p}_k}{\partial \kappa_j}
\end{equation}

Since the constraint applies to all points along the robot (proximal, distal, and tip points), the full constraint vector has one entry per point, and the Jacobian has dimensions $|\mathcal{X}_R| \times (m_p + m_d)$.

\subsection{Implementation Notes}

We compute all derivatives in a single forward pass through the kinematic chain. For each segment $i$:
\begin{enumerate}
\item Compute transformation $T_i$ and its derivative $\partial T_i / \partial \kappa_i$ using Eq.~\eqref{eq_constcurv_Tderiv}
\item Propagate previous gradients: $\partial T_{\text{prev}} / \partial \kappa_j \cdot T_i$ for $j < i$
\item Add new gradient: $T_{\text{prev}} \cdot \partial T_i / \partial \kappa_i$ for $j = i$
\item Extract point position derivatives from transformation derivatives using Eq.~\eqref{eq_point_derivative}
\end{enumerate}


\section*{Acknowledgment}

The authors would like to thank 
Dr. Michael G. Brandel (Department of Neurosurgery, University of California San Diego),
Dr. Jeremy J. Heit (Departments of Radiology and Neurosurgery, Stanford University), and
Dr. Alexander Norbash (School of Medicine, University of Missouri-Kansas City) for their clinical advice.
This work was supported in part by the National Institute of Health under R01 EB032417 and by the National Science Foundation under Grant 1944816. AM was also supported in part by the Powell Fellowship (Charles Lee Powell Foundation). The authors of this manuscript have patents on steering continuum robots in medical applications. 

\bibliography{biblio}

\end{document}